\pdfoutput=1

\documentclass[11pt,a4paper]{article}
\usepackage[hyperref]{emnlp2020}
\usepackage{times}
\usepackage{latexsym}

\usepackage{microtype}

\aclfinalcopy %
\def\aclpaperid{888} %

\usepackage{xcolor}
\definecolor{myblue}{rgb}{0.07, 0.04, 0.56}
\hypersetup{colorlinks=true,linkcolor=myblue,citecolor=myblue,urlcolor=myblue}

\usepackage{graphicx}
\usepackage{booktabs}
\usepackage{multirow}

\usepackage{enumitem}

\usepackage{soul}    %
\usepackage{xcolor}  %
\usepackage{tikz} %
\usepackage{pgfplots}
\usetikzlibrary{pgfplots.groupplots}
\pgfplotsset{compat=newest}
\usepgfplotslibrary{units}

\usepackage{algorithm}

\usepackage[disable]{todonotes}             %

\usepackage{textcomp} %

\usepackage{mathtools}  %

\usepackage{xspace}

\newcommand{\BJTParabanksysref}{Prism-ref w/ ParaBank 2 (Contrastive)} %
\newcommand{\BJTwikipluslm}{LM}  %
\newcommand{\BJTwikiplussysgivensrc}{Prism-src (This work)} %
\newcommand{\BJTwikiplussysref}{Prism-ref (This Work)} %
\newcommand{\BJTLASERrefLM}{LASER + LM (Contrastive)}
\newcommand{\BJTLASERref}{LASER}
\newcommand{\BJTmBARTsysref}{mBART (Contrastive)}
\newcommand{\sys}{\ensuremath{\mathrm{sys}}}
\newcommand{\reff}{\ensuremath{\mathrm{ref}}}
\newcommand{\src}{\ensuremath{\mathrm{src}}}

\newcommand{\BoldDenotes}{\textbf{Bold} denotes top scoring method and any other methods with whose 95\% confidence interval overlaps with that of a top method.}
\newcommand{\submission}{ {\text{\textdaggerdbl}} }
\newcommand{\qeasmetric}{ {\text{\textasteriskcentered}} }
\newcommand{\baseline}{ {\text{\textdagger}} }
\newcommand{\NoGu}{Our models were not trained on Gujarati (gu).}
\newcommand{\ensem}{} %
\newcommand{\nonen}{} %
\newcommand{\metric}[1]{\textsc{#1}}  %

\newcommand{\yisitwo}{\textsuperscript{a} \hspace{-1.9mm} }
\newcommand{\yisitwoText}{a:}
\newcommand{\yisitwosrl}{\textsuperscript{b} \hspace{-1.9mm} }
\newcommand{\yisitwosrlText}{b:}
\newcommand{\uni}{\textsuperscript{c} \hspace{-1.9mm} }
\newcommand{\uniText}{c:}
\newcommand{\uniplus}{\textsuperscript{d} \hspace{-1.9mm} }
\newcommand{\uniplusText}{d:}

\newcommand{\yisiboth}{\textsuperscript{a,b} \hspace{-3.5mm} }

\newcommand{\badres}[1]{  \textcolor{red}{\textit{{#1}}}  \hspace{-1.5mm} }
\newcommand{\badresNoOffset}[1]{   \hspace{-2mm}  \textcolor{red}{\textit{{#1}}}  \hspace{-2mm}  }

\usepackage{siunitx} %

\usepackage{fancyvrb}  %

\usepackage{glossaries}
\newacronym{nmt}{NMT}{Neural Machine Translation}
\newacronym{laser}{LASER}{Language-Agnostic SEntence Representations}
\newacronym{mt}{MT}{Machine Translation}
\newacronym{smt}{SMT}{Statistical Machine Translation}
\newacronym{dcce}{DCCE}{Dual Conditional Cross Entropy}
\newacronym{dp}{DP}{Dynamic Programming}
\newacronym{lstm}{LSTM}{Long Short-Term Memory}

\def\Snospace~{\S{}}

\usepackage{todonotes} %

\title{Automatic Machine Translation Evaluation in Many Languages\\via Zero-Shot Paraphrasing}

\author{Brian Thompson \\
  Johns Hopkins University \\
  \texttt{brian.thompson@jhu.edu} \\\And
  Matt Post \\
  Johns Hopkins University \\
  \texttt{post@cs.jhu.edu} \\}

\date{}

\begin{document}

\maketitle

\begin{abstract}

We frame the task of machine translation evaluation
as one of scoring machine translation output
with a sequence-to-sequence paraphraser,
conditioned on a human reference.
We propose training the paraphraser 
as a multilingual NMT system, treating paraphrasing
as a zero-shot translation task (e.g., Czech to Czech).
This results in the paraphraser's output mode being centered
around a copy of the input sequence,
which represents the best case scenario
where the MT system output matches a human reference.
Our method is simple and intuitive,
and does not require human judgements for training.
Our single model (trained in 39 languages)
outperforms or statistically ties with all prior metrics
on the WMT 2019 segment-level shared metrics task in all languages
(excluding Gujarati where the model had no training data).
We also explore using our model for the task of
quality estimation as a metric---conditioning on the source
instead of the reference---and find that it significantly outperforms
every submission to the WMT 2019 shared task on quality estimation
in every language pair.

\end{abstract}

\section{Introduction}

Machine Translation (MT) systems have improved dramatically in the past several years.
This is largely due to advances in neural MT (NMT) methods, %
but the pace of improvement would not have been possible without automatic MT metrics,
which provide immediate feedback on MT quality without
the time and expense
associated with obtaining human judgments of MT output.

\begin{figure}
 \centering
 \includegraphics[width=0.99\linewidth]{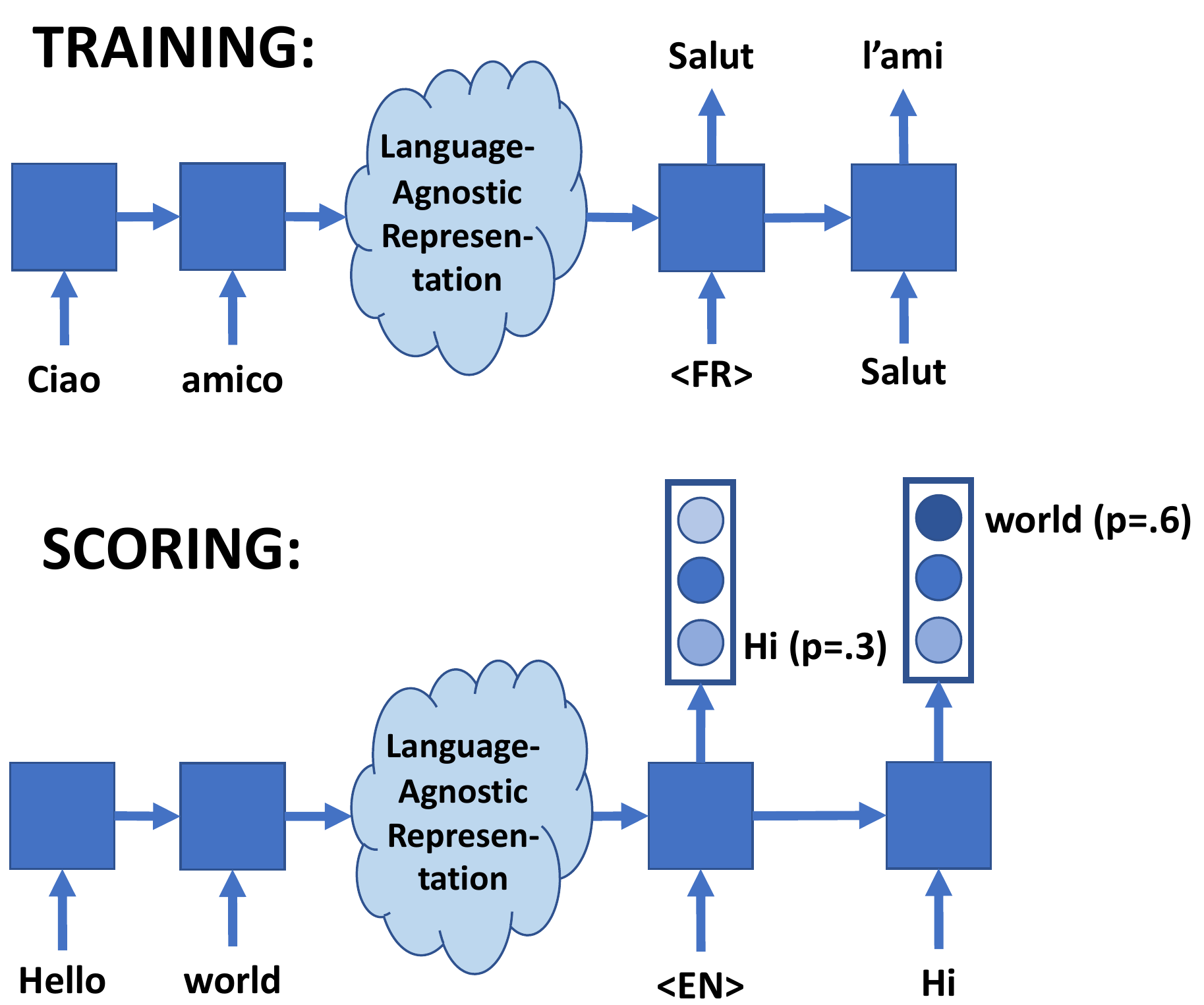}
 \caption{
   Our model is trained on multilingual
   parallel examples such as ``Ciao amico''
   translated to French is ``Salut l'ami.''
   At evaluation time, the model is used in zero-shot mode
   to score MT system outputs conditioned on their corresponding human references.
   For example, the MT system output ``Hi world''
   conditioned on the human reference
   ``Hello world'' is found to have token probabilities [0.3, 0.6].
}
 \label{fig:front_page}
\end{figure}

However, the improvements that existing automatic metrics helped enable
are now causing the correlation between human judgments and automatic metrics 
to break down \cite{ma-etal-2019-results, mathur-etal-2020-tangled}
especially for BLEU \cite{papineni-etal-2002-bleu},
which has been the de facto standard metric since
its introduction almost two decades ago.
The problem currently appears limited to very strong systems, but as hardware, modeling, and available training data
improve, it is likely BLEU will fail more frequently in the future.
This could prove extremely detrimental if the MT community
fails to adopt an improved metric,
as good ideas could quietly be discarded or rejected from publication
because they do not correlate with BLEU.
In fact, this may already be happening.

We propose using a
sentential, sequence-to-sequence
paraphraser to force-decode and score MT outputs
conditioned on their corresponding human references.
Our model implicitly represents the entire
(exponentially large) set of potential paraphrases
of a sentence, both valid and invalid;
by ``querying'' the model with a particular system output, 
we can use the model score to measure how well the system output paraphrases the human reference translation.
Our model is not trained on any human quality judgements, which are not available in many domains and/or language pairs.

The best possible MT output is one which perfectly matches a human reference;
therefore, for evaluation,
an ideal paraphraser would be one with an output distribution centered around a copy of its input sentence.
We denote such a model a ``lexically/syntactically unbiased paraphraser'' to distinguish
it from a standard paraphraser trained
to produce output which conveys the meaning of the input while also being \emph{lexically and/or syntactically different from it.}
For this reason, we propose using a multilingual NMT system as an unbiased paraphraser
by treating paraphrasing as zero-shot ``translation'' (e.g., Czech to Czech).
We show that a multilingual NMT model is much closer to an ideal lexically/syntactically unbiased paraphraser
than a generative paraphraser trained on synthetic paraphrases.
It also allows a single model to work in many languages,
and can be applied to the task of ``Quality estimation (QE) as a metric''
\cite{fonseca-etal-2019-findings} 
by conditioning on the source instead of the reference.
\autoref{fig:front_page} illustrates our method, which we denote \textit{Prism} (\underline{Pr}obability \underline{is} the \underline{m}etric).

We train a single model in 39 languages
and show that it:
\begin{itemize}[topsep=2pt, partopsep=2pt, itemsep=2pt,parsep=2pt]
    \item Outperforms or ties with 
    prior metrics
    and several contrastive neural methods
    on the segment-level WMT 2019 MT metrics task in every language pair;\footnote{Except for Gujarati, where we had no training data.}
    
\item Is able to discriminate between very strong neural systems at the system level, addressing a problem raised at WMT 2019; and
\item Significantly outperforms all QE metrics submitted to the WMT 2019 QE shared task 
\end{itemize}

\noindent Finally, we contrast the effectiveness of our model when scoring MT output using the source vs the human reference.
We observe that human references substantially improve performance,
and, crucially, allow our model to rank systems that are \emph{substantially better than our model at the task of translation.} 
This is important because it establishes that 
our method does not require building a state-of-the-art multilingual NMT model in order to produce a state-of-the-art MT metric capable of evaluating state-of-the-art MT systems.

We release our model, metrics toolkit, and preprocessed training 
data.\footnote{\url{https://github.com/thompsonb/prism}}

\section{Related Work}\label{sec:related}

\paragraph{MT Metrics}
Early MT metrics like BLEU \cite{papineni-etal-2002-bleu}
and NIST \cite{doddington2002automatic} use token-level n-gram overlap between the MT output and the human reference.
Overlap can also be measured at the character level \cite{popovic-2015-chrf,popovic-2017-chrf}
or using edit distance \cite{snover2006study}.
Many metrics 
use
word- and/or sentence-level embeddings,
including ReVal \cite{gupta-etal-2015-reval},
RUSE \cite{shimanaka-etal-2018-ruse},
WMDO \cite{chow-etal-2019-wmdo},
and ESIM \cite{mathur-etal-2019-putting}.
MEANT \cite{lo-wu-2011-meant} and MEANT 2.0 \cite{lo-2017-meant}
measure similarity 
between semantic frames and role fillers.
State-of-the-art methods including YiSi \cite{lo-2019-yisi}
and BERTscore \cite{bertscore_arxiv,bert-score}
rely on contextualized embeddings
\cite{devlin-etal-2019-bert} trained on large (non-parallel) corpora.
BLEURT \cite{bleurt} applies fine tuning of BERT, including training on prior human judgements.
In contrast, our work exploits parallel bitext and doesn't require training on human judgements.

\paragraph{Paraphrase Databases}%
Prior work explored using parallel bitext to identify 
phrase level paraphrases \cite{bannard-callison-burch-2005-paraphrasing,ganitkevitch-etal-2013-ppdb}
including bitext in multiple language pairs \cite{ganitkevitch-callison-burch-2014-multilingual}.
Paraphrase tables were, in turn,
used in MT metrics
to reward systems for paraphrasing
words \cite{banerjee-lavie-2005-meteor}
or phrases \cite{zhou-etal-2006-paraeval, denkowski-lavie-2010-extending}
from the human reference.
Our work can be viewed as extending this idea to the sentence level,
without having to enumerate the
millions or billions of paraphrases \cite{dreyer-marcu-2012-hyter}
for each sentence.

\begin{table*}[ht]
\footnotesize
\centering
\begin{tabular}{  @{\hskip0pt}  l  @{\hskip4pt}  l  @{\hskip4pt}  c   @{\hskip5pt}  c  @{\hskip5pt}  c  @{\hskip0pt} } 
\toprule 
     & \textbf{Word-level paraphraser log probabilities} & \textbf{H(out$|$in)} & \textbf{sBLEU} & \textbf{LASER} \\ 
\midrule 
\multirow{1}{*}{\begin{tabular}{@{\hskip0pt}l@{\hskip0pt}}Copy \\ \end{tabular}}  & \begin{tabular}{ @{\hskip0pt}  c  @{\hskip3pt}  c  @{\hskip3pt}  c  @{\hskip3pt}  c  @{\hskip3pt}  c  @{\hskip3pt}  c  @{\hskip3pt}  c  @{\hskip3pt}  c  @{\hskip3pt}  c  @{\hskip3pt}  c  @{\hskip3pt}  c @{\hskip0pt} } Jason & went & to & school & at & the & University & of & Madrid & . & $<$EOS$>$ \\ -0.08 & -0.26 & -0.16 & -0.16 & -0.12 & -0.11 & -0.14 & -0.10 & -0.10 & -0.11 & -0.10 \\ \end{tabular} & -0.13 & 100.0 & 1.000  \\ 
\midrule 
\multirow{1}{*}{\begin{tabular}{@{\hskip0pt}l@{\hskip0pt}}Disfluent \\ \end{tabular}}  & \begin{tabular}{ @{\hskip0pt}  c  @{\hskip3pt}  c  @{\hskip3pt}  c  @{\hskip3pt}  c  @{\hskip3pt}  c  @{\hskip3pt}  c  @{\hskip3pt}  c  @{\hskip3pt}  c  @{\hskip3pt}  c @{\hskip0pt} } Jason & went & \textbf{school} & at & \textbf{University} & of & Madrid & . & $<$EOS$>$ \\ -0.08 & -0.26 & \textbf{-7.21} & -0.12 & \textbf{-4.81} & -0.10 & -0.11 & -0.11 & -0.10 \\ \end{tabular} & -1.43 & 35.5 & 0.989  \\ 
\midrule 
\multirow{2}{*}{\begin{tabular}{@{\hskip0pt}l@{\hskip0pt}}Inadequate \\ \end{tabular}}  & \begin{tabular}{ @{\hskip0pt}  c  @{\hskip3pt}  c  @{\hskip3pt}  c  @{\hskip3pt}  c  @{\hskip3pt}  c  @{\hskip3pt}  c  @{\hskip3pt}  c  @{\hskip3pt}  c  @{\hskip3pt}  c  @{\hskip3pt}  c  @{\hskip3pt}  c  @{\hskip3pt}  c @{\hskip0pt} } Jason & \textbf{will} & \textbf{go} & to & school & at & the & University & of & Madrid & . & $<$EOS$>$ \\ -0.08 & \textbf{-9.77} & \textbf{-0.76} & -0.22 & -0.19 & -0.14 & -0.15 & -0.16 & -0.10 & -0.10 & -0.12 & -0.10 \\ \end{tabular} & -0.99 & 70.8 & 0.960  \\ 
\cmidrule{2-2} 
 & \begin{tabular}{ @{\hskip0pt}  c  @{\hskip3pt}  c  @{\hskip3pt}  c  @{\hskip3pt}  c  @{\hskip3pt}  c  @{\hskip3pt}  c  @{\hskip3pt}  c  @{\hskip3pt}  c  @{\hskip3pt}  c  @{\hskip3pt}  c  @{\hskip3pt}  c @{\hskip0pt} } Jason & went & to & school & at & the & University & of & \textbf{Berlin} & . & $<$EOS$>$ \\ -0.08 & -0.26 & -0.16 & -0.16 & -0.12 & -0.11 & -0.14 & -0.10 & \textbf{-10.34} & -0.12 & -0.10 \\ \end{tabular} & -1.06 & 78.3 & 0.957  \\ 
\midrule 
\begin{tabular}{@{\hskip0pt}l@{\hskip0pt}}Fluent \& \\ Adequate \\ \end{tabular}  & \begin{tabular}{ @{\hskip0pt}  c  @{\hskip3pt}  c  @{\hskip3pt}  c  @{\hskip3pt}  c  @{\hskip3pt}  c  @{\hskip3pt}  c  @{\hskip3pt}  c  @{\hskip3pt}  c @{\hskip0pt} } Jason & \textbf{attended} & the & University & of & Madrid & . & $<$EOS$>$ \\ -0.08 & \textbf{-2.01} & -1.63 & -0.42 & -0.10 & -0.09 & -0.16 & -0.10 \\ \end{tabular} & -0.57 & 41.1 & 0.918  \\ 
\bottomrule 
\end{tabular}
\caption{Example token-level log probabilities from our model for various output sentences, 
conditioned on input sentence (i.e., human reference) 
\textit{``Jason went to school at the University of Madrid.''}
H(out$|$in) denotes the average token-level log probability.
We observe that our model generally penalizes any deviations (\textbf{bolded}) from the input sentence,
but tends to penalize deviations which change the meaning 
of the sentence or introduce a disfluency more harshly 
than those which are fluent and adequate.
Sentence-level BLEU with smoothing=1 (``sBLEU'') 
and LASER embedding cosine similarity (``LASER'') are
shown for comparison. 
We note that LASER appears fairly insensitive to disfluencies, and sentenceBLEU struggles to reward valid paraphrases.
}\label{fig:mt_metric_sentscoreexampleEN} 
\end{table*} 

\paragraph{Multilingual NMT}
Multilingual NMT  \cite{dong-etal-2015-multi} has been shown
to rival performance of single language pair models 
in high-resource languages
\cite{aharoni-etal-2019-massively, arivazhagan2019massively}
while also improving low-resource translation via
transfer learning from higher-resource languages
\cite{zoph-etal-2016-transfer, nguyen-chiang-2017-transfer, neubig-hu-2018-rapid}.
An extreme low-resource setting is where
the system translates between languages seen during training,
but in a language \emph{pair}
where it did not see \textit{any} training data, denoted `zero-shot' translation.
Despite evidence that intermediate representations
are not truly language-agnostic \cite{kudugunta-etal-2019-investigating},
zero-shot translation has been shown successful,
especially between related languages \cite{johnson-etal-2017-googles, gu-etal-2018-universal, pham-etal-2019-improving}.

\paragraph{Generative Paraphrasing}%
Sentential paraphrasing can be accomplished 
by training an MT system on paraphrase examples instead of
translation pairs \cite{quirk-etal-2004-monolingual}.
While natural paraphrase datasets do exist
\cite{quirk-etal-2004-monolingual, coster-kauchak-2011-simple, fader-etal-2013-paraphrase, 10.1007/978-3-319-10602-1_48, federmann-etal-2019-multilingual},
they are somewhat limited.
An alternative is to start with much more plentiful bitext 
and back-translate one side into the language of the other
to create synthetic paraphrases on which to train 
\cite{prakash-etal-2016-neural, wieting-gimpel-2018-paranmt, hu-etal-2019-improved, parabank, hu-etal-2019-large}.
\citet{tiedemann-scherrer-2019-measuring} propose using
paraphrasing as a way to measure the semantic abstraction of multilingual NMT.
They also propose using a multilingual NMT model as a generative paraphraser.%
\footnote{We find that generating from a well trained multilingual NMT
system tends to produce copies of the input, as opposed to interesting paraphrases (see \autoref{appendix:genexamples}).}

\paragraph{Semantic Similarity}
Parallel corpora in many language pairs
have been used to produce fixed-size, multilingual sentence representations
\cite{schwenk-douze-2017-learning, wieting-etal-2017-learning, laser, wieting-etal-2019-simple, raganato-etal-2019-evaluation}.
LASER \cite{laser}, for example, trains a variant of NMT with a fixed-size intermediate
representation in 93 languages.
Embeddings produced by the encoder can be compared
to measure
intra- or inter-lingual semantic similarity.

\section{Method}\label{sec:method}

We propose using a paraphraser to force-decode and estimate probabilities of MT system
outputs, conditioned on their corresponding human references.
Let  $p(y_t | y_{i<t},x)$ be the probability our paraphraser
assigns to the $t$\textsuperscript{th} token in output sequence $y$, given
the previous output tokens $y_{i<t}$ and the input sequence $x$.
\autoref{fig:mt_metric_sentscoreexampleEN} shows an example of how token-level probabilities from our model
(described in \autoref{sec:experiments}) penalize
both fluency and adequacy errors given a human reference.
We consider two ways of combining token-level probabilities from the
model---sequence-level log probability ($G$) and average token-level log probability ($H$):
\begin{flalign}
G(y|x) &= \sum_{t=1}^{|y|} \log p(y_t | y_{i<t},x) \nonumber \\
H(y|x) &= \frac{1}{|y|} G(y|x) \nonumber
\end{flalign}
Let $\mathrm{sys}$ denote an MT system output, $\mathrm{ref}$ denote a human reference, and $\mathrm{src}$ denote the source.
We expect
scoring $\mathrm{sys}$ conditioned on $\mathrm{ref}$ to be most indicative of the quality of $\mathrm{sys}$.
However, we also explore scoring $\mathrm{ref}$ conditioned on $\mathrm{sys}$
as we find qualitatively that output sentences which drop
some meaning conveyed by the input sentence
are penalized less harshly by the model than
output sentences which contain extra information not present in the input.
Scoring in both directions
  to penalize the presence of information
  in one sentence but not the other is similar, in spirit,
  to methods which use bi-directional textual entailment
  as an MT metric \cite{pado-etal-2009-robust, khobragade2019machine}.%
\footnote{Conditional probabilities of MT systems in each direction have been shown effective at filtering MT training data \cite{junczys-dowmunt-2018-dual}.}

We postulate that the output sentence that best represents the meaning of an input sentence is,
in fact, simply a copy of the input sentence,
as precise word order and choice often convey subtle connotations.
As such, we seek a model 
whose output distribution is \emph{centered around a copy of the input sentence},
which we denote a ``lexically/syntactically unbiased paraphraser.''
While a standard generative paraphraser
is trained to retain semantic meaning, 
it does not meet our criteria because it is \emph{simultaneously}
trained to produce output which is lexically/syntactically different than its input,
a key element in generative paraphrasing \cite{bhagat-hovy-2013-squibs}.

We propose using a multilingual NMT system as a lexically/syntactically unbiased paraphraser.
A multilingual NMT system consists of an encoder which maps a sentence
in to an (ideally) language-agnostic semantic representation,
and decoder to map that representation back to a sentence.
The model has only seen bitext in training,
but we propose to treat paraphrasing as a zero-shot ``translation'' (e.g., Czech to Czech).

Because our model is multilingual,
we can also score MT system output
conditioned on the source sentence instead of the human reference.
This task is known as ``quality estimation (QE) as a metric,''
and was part of the WMT19 QE shared task \cite{fonseca-etal-2019-findings}.
We use ``Prism-ref'' to denote our reference-based metric
and ``Prism-src'' to denote our system applied as a QE metric.

Our final metric and QE metric are defined
based on results on our development set (see \autoref{sec:results:prelim}) as follows:
\begin{flalign}
\text{Prism-ref} &= \frac{1}{2} H(\mathrm{sys}|\mathrm{ref}) + \frac{1}{2} H(\mathrm{ref}|\mathrm{sys}) \nonumber \\
\text{Prism-src} &= H(\mathrm{sys}|\mathrm{src}) \nonumber
\end{flalign}
To obtain system-level scores, we average segment-level scores over all segments in the test set.%

\section{Experiments}\label{sec:experiments}

We train a multilingual NMT model and explore the extent to which
it functions as a lexically/syntactically unbiased paraphraser.
We then conduct several preliminary experiments on the WMT18 MT metrics data \cite{ma-etal-2018-results}
to determine how to best utilize the token-level probabilities
from the paraphraser,
and report results on the WMT19 system- and segment-level metric tasks \cite{ma-etal-2019-results} and QE as a metric task \cite{fonseca-etal-2019-findings}.

\subsection{Data Preparation}\label{subsec:dataprep}
Our method requires a model,
which in turn relies heavily on the data on which it is trained,
so we describe here the rationale behind
the design decisions made regarding the training data.
Full details sufficient for replication are provided in \autoref{appendix:datadetails}.
\paragraph{Language-Agnostic Representations}%
To encourage our intermediate representation to be
as language-agnostic as possible,
we choose datasets
with as much language pair diversity as possible
(i.e., not just en--* and *--en),
as \citet{kudugunta-etal-2019-investigating} %
has shown that encoder representation is affected by both the source language and target language.
While it is common to append the target language token to the source sentence,
we instead
prepend it to the target sentence so that
the encoder cannot
do anything target-language specific with this tag.
At test time, we force-decode the desired language tag
prior to scoring. 
\paragraph{Noise}
NMT systems are known to be sensitive to noise,
including sentence alignment errors \cite{khayrallah-koehn-2018-impact},
so we perform filtering with LASER \cite{schwenk-2018-filtering, chaudhary-etal-2019-low}.
We also perform language ID filtering using FastText \cite{joulin2016bag}
to avoid training the decoder with incorrect language tags.
\paragraph{Number of Languages}%
\citet{aharoni-etal-2019-massively}
found that performance of zero-shot translation
in a related language pair
increased substantially when increasing the number of languages
from 5 languages and 25,
with a performance plateau somewhere between 25 and 50 languages.
We view paraphrasing as zero-shot translation
between sentences in the same language, so
we expect to need a similar number of languages.

\paragraph{Copies}
We filter sentence pairs with excessive copies and partial copies,
as multiple studies \cite{ott2018analyzing,khayrallah-koehn-2018-impact}
have noted that MT performance degrades substantially 
when systems are exposed to copies in training.

\subsection{Model Training}

We train a Transformer \cite{vaswani2017attention} model 
with approximately 745M parameters to translate between 39 languages. 
The full list of languages and data amounts used is provided in \autoref{appendix:datadetails},
and model training details sufficient for replication are given in \autoref{appendix:training}.
Training a single large model
consumed the majority of our compute budget,
thus performing ablations 
is beyond the scope of this work.

Our data comes primarily from WikiMatrix
\cite{DBLP:journals/corr/abs-1907-05791},
Global Voices,\footnote{\url{http://casmacat.eu/corpus/global-voices.html}}
EuroParl \cite{koehn2005europarl},
SETimes,\footnote{\url{http://nlp.ffzg.hr/resources/corpora/setimes/}}
and 
United Nations \cite{eisele2010multiun}.
The data processing described above and in \autoref{appendix:datadetails} 
results in 99.8M sentence pairs in 39 languages.%
\footnote{For every sentence pair (a,b) in our 99.8M examples, we train on both (a,b) and (b,a)}
The most common language is English, at 16.7\% of our data, while 
the least common 20 languages account for 21.9\%.

\subsection{Baselines and Contrastive Methods}
We compare to all systems from the WMT19
shared metrics task,
as well as BERTscore \cite{bert-score}
and the recent BLEURT method \cite{bleurt}.
We also explore several contrastive methods. %
Training details sufficient for replication for each model/baseline are given in \autoref{appendix:training}.
\paragraph{Generative Sentential Paraphraser}%
We compare scoring with our Prism model vs
a standard, English-only paraphraser trained on the ParaBank 2 dataset 
\cite{hu-etal-2019-large}. ParaBank 2 contains $\sim$ 50M synthetic paraphrastic pairs derived from back-translating a Czech--English corpus, and the authors report state-of-the-art paraphrasing results.
\paragraph{Auto-encoder}
Auto-encoders provide an alternative means of training seq2seq models, 
without the need for parallel bitext.
We compare to scoring with the ``multilingual denoising pre-trained model'' (mBART) 
of \citet{liu-etal-2020-multilingual}, as it works in all languages of interest.
\paragraph{LASER}
We explore using the cosine distance
between LASER embeddings of the MT output and human reference,
using the pretrained 93-language model
provided by the authors.\footnote{\url{https://github.com/facebookresearch/LASER}}
We are particularly interested in LASER
as it, like our model, is trained on
parallel bitext in many languages.
\paragraph{Language Model}
We find qualitatively that LASER is fairly insensitive to disfluencies (see \autoref{fig:mt_metric_sentscoreexampleEN}),
so we also explore augmenting it with language model (LM) scores of the system outputs.
We train a multilingual language model (see \autoref{appendix:training}) on the same data as our multilingual NMT system.
\subsection{Paraphraser Bias}
We expect that a lexically/syntactically unbiased measure of translation quality should (on average)
increase with increased lexical similarity between a translation and reference.
To explore the extent to which Prism and the model trained on ParaBank 2 are biased,
we consider average $H(\sys|\reff)$ as a function of binned lexical similarity (approximated by sentBLEU, with smoothing=1)
for all $(\sys, \reff)$ pairs for all systems submitted to WMT19 in all language pairs into English.
We also contrast the conditional probabilities of three outputs for the same input:
(1) the sequence generated by the model via beam search;
(2) a copy of the input; and
(3) a human paraphrase of the input.
Finally, we generate from the model using beam search and examine the outputs to see how much they differ from the inputs.

\subsection{MT Metrics Evaluation}
We report results and statistical significance using scripts released with the WMT19 shared task.
Segment-level performance is reported as the Kendall's $\tau$ variant
used in the shared task, and system-level performance is reported
as Pearson correlation
with the mean of the human judgments. 
Bootstrap resampling
\cite{koehn-2004-statistical, graham-etal-2014-randomized}
is used to estimate confidence intervals for each metric, and
metrics with non-overlapping
95\% confidence intervals are identified as having a
statistically significant difference in performance.

\section{Results}

\begin{figure*}
 \centering
 \includegraphics[width=0.75\linewidth]{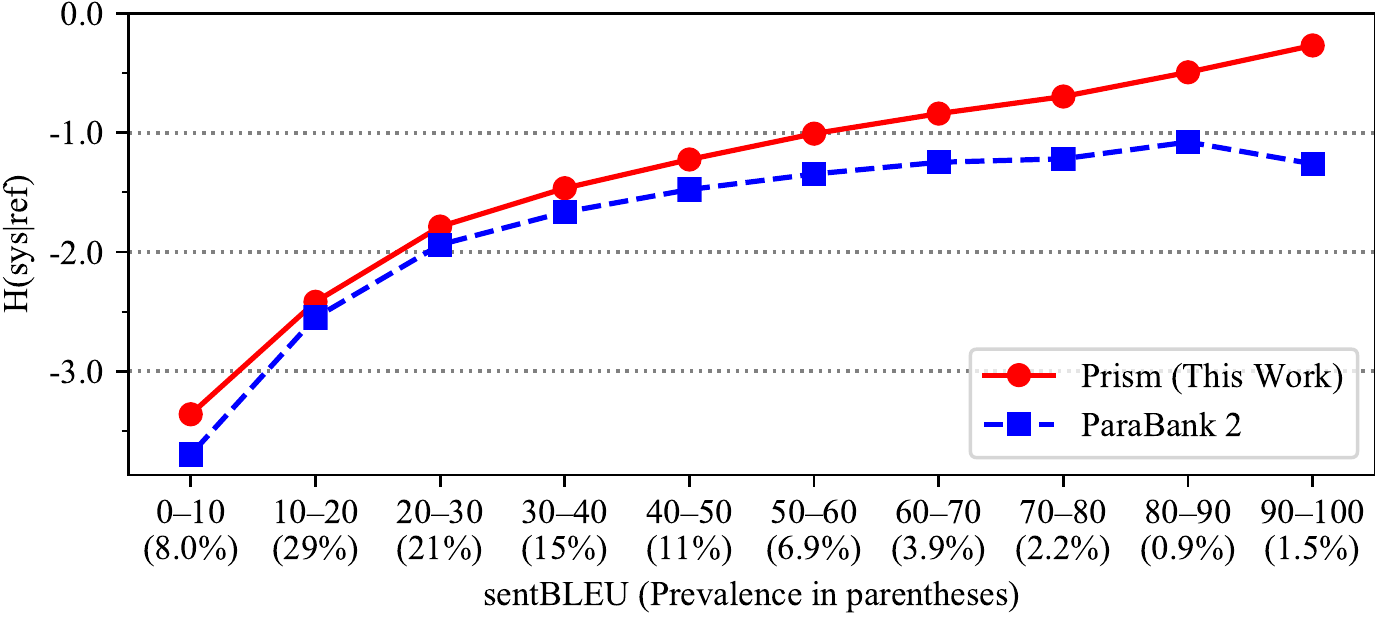}
 \caption{Average $H(\sys|\reff)$ as a function of average lexical difference  (as measured by sentBLEU)
   for every English $(\sys, \reff)$ pair submitted to WMT19, for both the Prism and ParaBank 2 paraphrasers.
    $(sys,ref)$ pairs are split into 10 sentBLEU bins of uniform width. Fraction of total data in each bin is shown on x-axis (in parentheses). 
    }
 \label{fig:bias_plot}
\end{figure*}

\begin{table*}[h!]
\centering 
\footnotesize
    \addtolength{\tabcolsep}{-2pt} 
  \begin{tabular}{@{\hskip 0pt}lrrrrrrrr|rrr@{\hskip 0pt}}

\toprule
                                                                                                          & {\bf en--cs}     & {\bf en--de}    & {\bf en--fi}    & {\bf en--gu} & {\bf en--kk} & {\bf en--lt}    & {\bf en--ru}    & {\bf en--zh}  & {\bf de--cs}  & {\bf de--fr} & {\bf fr--de}    \\
\midrule                                                                                                                                                                                                                                                                                 
\nonen \metric{BERTscore}      \cite{bert-score}                                           &       0.485     &      0.345     &      0.524     & {\bf 0.558} & {\bf 0.533} &      0.463     &      0.580     &      0.347   &    0.352  &       0.325 &       0.274       \\
\nonen \metric{EED}${}^\submission$                \cite{stanchev-etal-2019-eed}                           &       0.431     &      0.315     &      0.508     & {\bf 0.568} &      0.518  &      0.425     &      0.546     &      0.257   &    0.345  &       0.301 &       0.267       \\
\nonen \metric{YiSi-1}${}^\submission$             \cite{lo-2019-yisi}                                     &       0.475     &      0.351     &      0.537     & {\bf 0.551} & {\bf 0.546} &      0.470     &      0.585     & {\bf 0.355}  &    0.376  &       0.349 &       0.310       \\
\nonen \metric{YiSi-1\_srl}${}^\submission$        \cite{lo-2019-yisi}                                     &          $-$    &      0.368     &         $-$    &         $-$ &         $-$ &         $-$    &         $-$    & {\bf 0.361}  &     $-$   &       $-$   &       0.299       \\
\midrule
\nonen \BJTwikiplussysref                                                                                  & {\bf 0.582}     & {\bf 0.427}    & {\bf 0.591}    &     0.313  & {\bf 0.531} & {\bf 0.558}    &      0.584     & {\bf 0.376}  & {\bf 0.458} & {\bf 0.453} & {\bf 0.426} \\
\nonen \BJTLASERrefLM                                                                                      &      0.535      &      0.401     &      0.568     &     0.306  &      0.408  &      0.503     & {\bf 0.640}    & {\bf 0.356}  &      0.431  &      0.401  & {\bf 0.381} \\
\nonen \BJTmBARTsysref                                                                                      &      0.345      &      0.302     &      0.401     &      0.528  &      0.462  &      0.365     &      0.443     &      0.280  &      0.262  &      0.255  &      0.236 \\
\bottomrule
\end{tabular}
\begin{tabular}{@{\hskip 0pt}lrrrrrrr@{\hskip 0pt}}
\toprule
                                                                                                           & {\bf de--en}     & {\bf fi--en} & {\bf gu--en}    & {\bf kk--en} & {\bf lt--en} & {\bf ru--en}    & {\bf zh--en} \\
\midrule
\nonen \metric{BERTscore}      \cite{bert-score}                                                           &       0.176     &       0.345 & {\bf 0.320}    & {\bf 0.432} & {\bf 0.381} & {\bf 0.223}    & {\bf 0.430} \\
\nonen \metric{BLEURT}         \cite{bleurt}                                                               & {\bf 0.204}     & {\bf 0.367} & {\bf 0.311}    & {\bf 0.447} & {\bf 0.387} & {\bf 0.228}    & {\bf 0.423} \\
\nonen \metric{ESIM}${}^\submission$               \cite{chen-etal-2017-enhanced,mathur-etal-2019-putting} &       0.167     &      0.337  &      0.303     & {\bf 0.435} &      0.359  &      0.201     &      0.396  \\
\nonen \metric{YiSi-1}${}^\submission$             \cite{lo-2019-yisi}                                     &       0.164     &      0.347  & {\bf 0.312}    & {\bf 0.440} & {\bf 0.376} & {\bf 0.217}    & {\bf 0.426} \\
\nonen \metric{YiSi-1\_srl}${}^\submission$        \cite{lo-2019-yisi}                                     &  {\bf 0.199}    &      0.346  &      0.306     & {\bf 0.442} & {\bf 0.380} & {\bf 0.222}    & {\bf 0.431} \\
\midrule
\nonen \BJTwikiplussysref                                                                                  & {\bf 0.204}     & {\bf 0.357} & {\bf 0.313}    & {\bf 0.434} & {\bf 0.382} & {\bf 0.225}    & {\bf 0.438} \\
\nonen \BJTParabanksysref                                                                                  &      0.184      & {\bf 0.341} & {\bf 0.326}    &      0.425  & {\bf 0.373} &      0.207     & {\bf 0.432} \\
\nonen \BJTLASERrefLM                                                                                      &      0.190      &      0.335  & {\bf 0.319}    &       0.428  & {\bf 0.368} &      0.207     &      0.416  \\
\nonen \BJTmBARTsysref                                                                                      &      0.136      &      0.255  &      0.246     &      0.377  &      0.298  &      0.162     &      0.349  \\
\bottomrule
\end{tabular}\caption{WMT19 segment-level human correlation ($\tau$), to non-English (top) and to English (bottom).
  \BoldDenotes{ }
  $\submission$:WMT19 Metric Submission.
  For brevity, only competitive baselines are shown. For complete results see \autoref{appendix:wmt19_seg}. \NoGu{ }
``LASER + LM'' denotes the optimal linear combination found on the development set.}\label{fig:mt_metric_wmt19_Segment_fromEnglish_fubar}
\end{table*}

\subsection{Paraphraser Bias Results}\label{sec:bias_results}

We find $H(\sys|\reff)$ increases monotonically with sentBLEU for the Prism model, 
but the model trained on ParaBank 2 has nearly the same scores for
output with sentBLEU in the range of 60 to 100;
however that range accounts for only about 8.5\% of all system outputs (see \autoref{fig:bias_plot}). 
We find that a copy of the input is almost as probable as beam search output for the Prism model.
In contrast, the model trained on ParaBank 2 prefers its own beam search output to a copy of the input.
Additionally, beam search from our model produces output which
is more lexically similar to the input (BLEU of 82.8 with respect to input, vs 31.9 for ParaBank 2).
ParaBank 2 tends to change the output in ways which occasionally significantly alter the meaning of the sentence.
See \autoref{appendix:genexamples} for more details.
All of these findings support our hypothesis that our model is closer 
to an ideal lexically/syntactically unbiased paraphraser
than the contrastive model trained on synthetic paraphrases.

\subsection{Preliminary (Development) Results}\label{sec:results:prelim}

\begin{table*}
\centering 
\footnotesize
    \addtolength{\tabcolsep}{-2pt} 
\begin{tabular}{lrrrrrrrr|rrr}
\toprule
{}                                                                 & {\bf en--cs}     & {\bf en--de}    & {\bf en--fi}    & {\bf en--gu} & {\bf en--kk} & {\bf en--lt}    & {\bf en--ru}    & {\bf en--zh}  & {\bf de--cs}  & {\bf de--fr} & {\bf fr--de}    \\
\midrule
\nonen \metric{BERTscore}      \cite{bert-score}                   &   0.868 &    \badres{-0.722}  &  0.859 &  0.922 &  0.288 &  0.955 & 0.953 & 0.982 &  0.976 &   0.707 &  0.973  \\
\nonen \metric{BLEU}${}^\baseline$   \cite{papineni-etal-2002-bleu}  &   0.930 &    \badres{-0.370}  &  0.898 &  0.860 &  0.181 &  0.925 & 0.753 & 0.987 &  0.812 &   0.495 &  0.983  \\
\nonen \metric{YiSi-1}${}^\submission$    \cite{lo-2019-yisi}        &   0.847 &     \badres{-0.220}  & 0.976 &  0.917  & 0.342 &  0.838 & 0.963  & 0.990 &  0.967 &  0.677 &   0.967 \\
\nonen \metric{YiSi-1\_srl}${}^\submission$   \cite{lo-2019-yisi}    &    $-$  &     \badres{-0.378}  &  $-$  &   $-$   &   $-$ &   $-$  &  $-$   & 0.994 &   $-$  &   $-$  &   0.974 \\
\midrule
\BJTwikiplussysref                                                 &   0.952 &               0.278    &  0.886 &  0.863 &  0.693 &           0.862 & 0.975 & 0.966 &  0.968 &   0.648  & 0.998   \\
\BJTLASERrefLM                                                     &  0.961  &               0.377    &  0.903 &  0.509 &  0.605 &           0.743 & 0.962 & 0.985 &  0.947 &   0.774  & 0.975  \\
\nonen \BJTmBARTsysref                                             &  0.936  &    \badres{-0.834}   & 0.966  &  0.912 &  0.224 &         0.946   & 0.968 & 0.986 &  0.964 &   0.944  & 0.874 \\
\bottomrule
\end{tabular}
\begin{tabular}{lrrrrrrr}
\toprule
{}                                                                       & {\bf de--en}     & {\bf fi--en} & {\bf gu--en}    & {\bf kk--en} & {\bf lt--en} & {\bf ru--en}    & {\bf zh--en} \\
\midrule

\nonen \metric{BERTscore}           \cite{bert-score}                    &       0.272     &  0.683          &  0.913    &  0.897  &  0.753    &  0.456           &     \badres{-0.220}  \\
\nonen \metric{BLEU}${}^\baseline$    \cite{papineni-etal-2002-bleu}       &    \badres{-0.822} &    \badres{-0.275} &   0.966   &  0.958  &  0.625    &     \badres{-0.356} &    \badres{-0.694}   \\
\nonen \metric{BLEURT}      \cite{bleurt}                                &       0.953     &  0.714          & 0.881     &  0.929  &  0.841    &  0.522  &      0.660  \\
\nonen \metric{YiSi-1}${}^\submission$        \cite{lo-2019-yisi}          &       0.045     &  0.610    &  0.962    &  0.887 &  0.552    &  0.365    &        \badres{-0.067}  \\
\nonen \metric{YiSi-1\_srl}${}^\submission$   \cite{lo-2019-yisi}          &       0.081     &  0.580    &  0.959    &  0.874 &  0.560    &  0.342    &        \badres{-0.069}  \\
\midrule
\nonen \BJTwikiplussysref                                                &       0.401     &   0.719   &   0.896   &  0.796  &  0.877    & 0.431             &           0.523   \\
\nonen \BJTLASERrefLM                                                    &       0.957     &   0.768   &   0.867   &  0.870  &  0.615    &  0.596            &           0.733   \\
\nonen \BJTmBARTsysref                                                   &  \badres{-0.739}  & 0.559  &  0.913     &  0.902 &   0.491     &  \badres{-0.103}  & \badres{-0.295}  \\
\bottomrule
\end{tabular}%
\caption{WMT19 system-level human correlation (Pearson), for top 4 systems only, 
to non-English (top) and to English (bottom), for selected metrics.
Negative correlations with human judgments shown in \badresNoOffset{red} for emphasis.
$\baseline$:WMT19 Baseline 
$\submission$:WMT19 Metric Submission.
``LASER + LM'' denotes the optimal linear combination found on the development set.
\NoGu}\label{fig:mt_metric_top4}
\end{table*}

\begin{table*}[h!]
\centering
  \footnotesize
    \addtolength{\tabcolsep}{-2pt} 
\begin{tabular}{lrrrrrrrr|rrr}
\toprule
                            & {\bf en--cs}    & {\bf en--de}    & {\bf en--fi} & {\bf en--gu} & {\bf en--kk} & {\bf en--lt} & {\bf en--ru}   & {\bf en--zh} { }   & { }{\bf de--cs} & {\bf de--fr} &    {\bf fr--de} \\
\midrule
Best WMT19 QE as Metric     & 0.069\yisitwo       & 0.236\yisitwosrl    &     0.351\uni   &     0.147\yisitwo   &      0.187\yisitwo   &      0.003\yisitwo  &       0.226\uni  &     0.044\yisitwo  { }  & { }     0.199\yisitwo  &      0.186\yisitwo  &      0.066\yisitwo     \\
\BJTwikiplussysgivensrc     & {\bf 0.470}         & {\bf 0.402}         & {\bf 0.555}     &  {\bf 0.215}        & {\bf 0.507}          & {\bf 0.499}         & {\bf 0.486}      & {\bf 0.287}        { }  & { } {\bf 0.444} & {\bf 0.371} &    {\bf 0.316} \\
\bottomrule
\end{tabular}
\begin{tabular}{lrrrrrrr}
\toprule
                           & {\bf de--en}{ }        & {\bf fi--en} & {\bf gu--en} & {\bf kk--en} & {\bf lt--en} & {\bf ru--en} & {\bf zh--en} \\
\midrule
Best WMT19 QE as Metric    &      0.068\yisiboth { }         &     0.211\uniplus   &    $-$0.001\yisitwo &      0.096\yisitwo  &      0.075\yisitwo  &      0.089\uniplus  &      0.253\yisitwo  \\
\BJTwikiplussysgivensrc    & {\bf 0.109} { }        & {\bf 0.300} & {\bf 0.102} & {\bf 0.391} & {\bf 0.356} & {\bf 0.178} & {\bf 0.336} \\
\bottomrule
\end{tabular}\caption{WMT19 segment-level human correlation ($\tau$) for QE as Metric systems (which have access to the source only, not the reference). 
  \BoldDenotes{ } \NoGu {}
  For brevity, only the best QE-metric for each language pair is shown---for full results see \autoref{appendix:wmt19_sys}.
  \yisitwoText \metric{YiSi-2} \cite{lo-2019-yisi}
  \yisitwosrlText \metric{YiSi-2\_srl} \cite{lo-2019-yisi}
  \uniText \metric{UNI} \cite{yankovskaya-etal-2019-quality}
  \uniplusText \metric{UNI+} \cite{yankovskaya-etal-2019-quality}.
}\label{fig:mt_metric_wmt19_segment_qe}
\end{table*}

We find that length-normalized log probability ($H$) slightly outperforms un-normalized log probability ($G$).
When using the reference, we find an equal weighting of
$H(\sys|\reff)$ and $H(\reff|\sys)$ to be approximately optimal,
but we find that when using the source, $H(\src|\sys)$ does not appear to add
useful information to $H(\sys|\src)$. 
Full results can be found in \autoref{appendix:prelimresults}.
These findings were used to select the 
Prism-ref and Prism-src definitions (\autoref{sec:method}).

We find that the
probability of $\sys$ as estimated by an LM,
as well as and the cosine distance between LASER embeddings of $\sys$ and $\reff$, 
both have decent correlation with human judgments and are complementary.
However, cosine distance between LASER 
embeddings of $\sys$ and $\src$ 
have only weak correlation.  %
  
\subsection{Segment-Level Metric Results}

Segment-level metric results are shown in \autoref{fig:mt_metric_wmt19_Segment_fromEnglish_fubar}.
On language pairs into non-English, 
we outperform prior work by a statistically significant margin in 7 of 11 language pairs\footnote{In en--ru, Prism-ref is statistically tied with YiSi-1, ESIM, and BERTscore.}
and are statistically tied for best in the rest,
with the exception of Gujarati (gu) where the model had no training data.
Into English, our metric is statistically tied with the best prior work in every language pair.
Our metric tends to significantly outperform our contrastive LASER + LM and mBART methods,
although LASER + LM performs surprisingly well in en--ru.

\subsection{System-Level Metric Results}

\autoref{fig:mt_metric_top4} shows system-level metric performance
on the top four systems submitted to WMT19
compared to selected metrics.
While correlations %
are not high 
in all cases for Prism, they are at least all positive. 
In contrast, BLEU has negative correlation in 5 language pairs, 
and BERTscore and YiSi-1 variants are each negative in at least two.
BLEURT has positive correlations in all language pairs into English,
but is English-only.
Note that Pearson's correlation coefficient may be unstable in this setting \cite{mathur-etal-2020-tangled}.
For full top four system-level results see \autoref{appendix:wmt19_top4}.

We do not find the system-level results computed against \emph{all} submitted MT systems (see \autoref{appendix:wmt19_sys}) to be particularly interesting; 
as noted by \citet{ma-etal-2019-results}, 
a single weak system can result in
high overall system-level correlation even for a very poor metric.

\subsection{QE as a Metric Results}

We find that our reference-less Prism-src 
outperforms all QE as a metrics systems from the WMT19 shared task
by a statistically significant margin,
in every language pair at segment-level human correlation (\autoref{fig:mt_metric_wmt19_segment_qe}),
and outperforms or statistically ties at system-level human correlation (\autoref{appendix:wmt19_sys}).

\section{Analysis and Discussion}\label{vslaser}

\paragraph{How helpful are human references?}%
The fact that our model is multilingual
allows us to explore the extent to which
the human reference actually
improves our model's ability to judge MT system output,
compared to using the source instead.
The underlying assumption with any MT metric
is that the work done by the human translator makes
it easier to automatically judge the quality of MT output.
However, if our model
or the MT systems being judged were strong enough,
we would expect this assumption to break down.

\begin{table}
  \normalsize
  \centering
\begin{tabular}{lrrr}
\toprule
Lang   & \multicolumn{3}{c}{BLEU}   \\ 
Pair  & WMT19 Best & Multilingual & $\Delta$ \\ 
\midrule
de--cs  & 20.1\textdagger                     & 21.8\textcolor{white}{\textdaggerdbl} & +1.7 \\ 
de--en  & 42.8\textcolor{white}{\textdagger}  & 35.5\textcolor{white}{\textdaggerdbl} & -7.3 \\ 
de--fr  & 37.3\textcolor{white}{\textdagger}  & 33.9\textcolor{white}{\textdaggerdbl} & -3.4 \\
en--cs  & 29.9\textcolor{white}{\textdagger}  & 24.2\textcolor{white}{\textdaggerdbl} & -5.7 \\ 
en--de  & 44.9\textcolor{white}{\textdagger}  & 38.1\textcolor{white}{\textdaggerdbl} & -6.8 \\ 
en--fi  & 27.4\textcolor{white}{\textdagger}  & 21.9\textcolor{white}{\textdaggerdbl} & -5.5 \\ 
en--gu  & 28.2\textcolor{white}{\textdagger}  & 0.0\textdaggerdbl                     & -28.2 \\ 
en--kk  & 11.1\textcolor{white}{\textdagger}  & 8.6\textcolor{white}{\textdaggerdbl}  & -2.5 \\ 
en--lt  & 20.1\textcolor{white}{\textdagger}  & 15.0\textcolor{white}{\textdaggerdbl} & -5.1 \\ 
en--ru  & 36.3\textcolor{white}{\textdagger}  & 28.1\textcolor{white}{\textdaggerdbl} & -8.2 \\ 
en--zh  & 44.6\textcolor{white}{\textdagger}  & 30.1\textcolor{white}{\textdaggerdbl} & -14.5 \\ 
fi--en  & 33.0\textcolor{white}{\textdagger}  & 26.2\textcolor{white}{\textdaggerdbl} & -6.8 \\ 
fr--de  & 35.0\textcolor{white}{\textdagger}  & 26.4\textcolor{white}{\textdaggerdbl} & -8.6 \\ 
gu--en  & 24.9\textcolor{white}{\textdagger}  & 0.4\textdaggerdbl                     & -24.5 \\ 
kk--en  & 30.5\textcolor{white}{\textdagger}  & 27.7\textcolor{white}{\textdaggerdbl} & -2.8 \\ 
lt--en  & 36.3\textcolor{white}{\textdagger}  & 28.5\textcolor{white}{\textdaggerdbl} & -7.8 \\ 
ru--en  & 40.1\textcolor{white}{\textdagger}  & 36.1\textcolor{white}{\textdaggerdbl} & -4.0 \\ 
zh--en  & 39.9\textcolor{white}{\textdagger}  & 20.6\textcolor{white}{\textdaggerdbl} & -19.3 \\ 
\bottomrule
\end{tabular}
\caption[WMT19 BLEU Scores vs Best WMT Systems]{BLEU scores for our multilingual NMT system on WMT19 testsets, compared to best system from WMT19. Our multilingual system achieves state-of-the-art performance as an MT metric despite substantially under performing all the best WMT19 MT systems at translation (excluding unsupervised). \textdagger: WMT systems were unsupervised (no parallel data). \textdaggerdbl: Multilingual system did not train on Gujarati (gu). Systems are not trained on the same data, so this should not be interpreted as a comparison between multilingual and single-language pair MT. ISO~639-1 language codes.}\label{fig:mt_metric_wmt19bleuplot}
\end{table}

Comparing the performance of our method with access to the human reference (Prism-ref)
vs our method with access to only the source (Prism-src),
we find that the reference-based method statistically
outperforms the source-based method in all but one language pair.
We find the case where they are not statistically different, de--cs, to be particularly interesting:
de--cs was the only language pair in WMT19 where the systems were unsupervised (i.e., did not use parallel training data).
As a result, it is the only language pair where our model outperformed
the best WMT system at translation. In most cases, our model is \emph{substantially worse at translation} than the best WMT systems.
For example, in en--de and zh--en, two language pairs where strong NMT systems were especially problematic for MT metrics, %
the Prism model is 6.8 and 19.2 BLEU points behind the strongest WMT systems, respectively
(see \autoref{fig:mt_metric_wmt19bleuplot} for the Prism model compared to the best system submitted in each WMT19 language pair).
Thus the performance difference between Prism-ref and Prism-src
would suggest that
the model needs no help in judging MT systems which are weaker than it is, but
\emph{the human references are assisting our model in evaluating MT systems which are stronger than it is.} 
This means that we have not simply
reduced the task of MT evaluation to that of building a state-of-the-art MT system.
We see that a good (but not state-of-the-art) multilingual NMT system
can be a state-of-the-art MT metric and judge state-of-the-art MT systems.

Finally, with the exception of de--cs discussed above,
we see statistically significant improvements for Prism-ref over Prism-src
both into English (where human judgments were reference-based)
and into non-English (where human judgments were source-based).
This suggests that the high correlation of Prism-ref with human judgements
is not simply the result of reference bias \cite{fomicheva-specia-2016-reference}.

\paragraph{Does paraphraser bias matter?}
Our lexically/syntactically unbiased paraphraser tends to outperforms the generative English-only ParaBank 2 paraphraser, but usually not by a statistically significant margin. %
Analysis indicate the lexical/syntactic bias is only harmful in somewhat infrequent cases where MT systems match or nearly match the reference,
suggesting it would be more detrimental with stronger systems or multiple references.
Our multilingual training method is much simpler than 
the alternative of creating synthetic paraphrases and training
individual models in 39 languages, 
and our model may benefit from transfer learning to lower-resource languages. 

\paragraph{Does fluency matter?}
Despite NMT being very fluent,
our results %
suggest that fluency is fairly discriminative,
especially in non-English:
LM scoring outperforms sentenceBLEU at segment-level correlation
in 7/10 language pairs to non-English languages (excluding Gujarati), for example.
This is consistent with recent findings that LM scores can be used to augment BLEU \cite{edunov-etal-2020-evaluation}.

\paragraph{Can we measure adequacy and fluency separately?}
The proposed method
significantly outperforms the contrastive LASER-based method in most language pairs, even when LASER is augmented with a language model.
This suggests that 
jointly optimizing a model for adequacy and fluency 
is better than optimizing them independently and combining after the fact---this is unsurprising given that neural MT
has shown significant improvements over statistical MT,
where a phrase table and language model were trained separately. 

\paragraph{Can we train on monolingual data instead of bitext?}
The proposed method significantly outperforms scoring with the mBART auto-encoder,
which is trained on large amounts of monolingual data,
despite using substantially less compute power 
(1.3 weeks on 8 V100s for Prism vs 2.5 weeks on 256 V100s for mBART).  %

\section{Conclusion and Future Work}
We show that a multilingual NMT system
can be used as a lexically/syntactically unbiased, multilingual paraphraser,
and that the resulting paraphraser
can be used as an MT metric and QE metric.
Our method achieves state-of-the-art performance on the most recent WMT shared metrics and QE tasks, 
without training on prior human judgements.

We release a single model which supports 39 languages.
To the best of our knowledge, we are the first to release a large multilingual NMT system, and we hope others follow suit.
We are optimistic our method will improve further  
as stronger multilingual NMT models become publicly available.

We compare our method to several contrastive methods and
present analysis showing that we have not simply 
reduced the task of evaluation to that of building a state-of-the-art MT system;
the work done by the human translator to create references
helps the evaluation model to
judge systems that are stronger (at translation) than it is.

Nothing in our method is specific to sentence-level MT. In future work,
we would like to extend Prism to paragraph- or document-level evaluation
by training a paragraph- or document-level multilingual NMT system,
as there is growing evidence that MT evaluation would be better conducted at the document level,
rather than the sentence level \cite{laubli-etal-2018-machine}.

\section*{Acknowledgments}
Brian Thompson is supported 
by the National Defense Science and Engineering Graduate (NDSEG) Fellowship.

\bibliography{anthology,emnlp2020,thesis}
\bibliographystyle{acl_natbib}

\onecolumn
\appendix

\clearpage

\section{Generation Examples}\label{appendix:genexamples}

\autoref{tab:genexamples} shows sentences generated from both our model and the model trained on ParaBank 2.

We also contrast the conditional probabilities of three outputs for the same input:
(1) the sequence generated by the model via beam search;
(2) a copy of the input; and (3) a human paraphrase of the input.
We use the English side of the zh--en newstest17 \cite{bojar-etal-2017-findings}
as input, so that we can use the second human reference released by \citet{DBLP:journals/corr/abs-1803-05567} 
as a human paraphrase.
\autoref{fig:paraphraser_input-bias} shows the results of scoring a copy of the input, a human paraphrase of the input, and a model's beam search output, for both our multilingual paraphraser and the ParaBank 2 model.

\begin{figure*}
\footnotesize
\begin{tabular}{p{0.12\linewidth}p{0.8\linewidth}}
\toprule
REFERENCE & 28-Year-Old Chef Found Dead at San Francisco Mall\\
THIS WORK & 28-Year-Old Chef Found Dead at San Francisco Mall\\
PARABANK 2 & 28-year-\textbf{o}ld \textbf{c}hef \textbf{f}ound \textbf{d}ead \textbf{in a mall in} San Francisco\\
\midrule
REFERENCE & A 28-year-old chef who had recently moved to San Francisco was found dead in the stairwell of a local mall this week.\\
THIS WORK & A 28-year-old chef who had recently moved to San Francisco was found dead in the stairwell of a local mall this week.\\
PARABANK 2 & \textbf{Earlier this week,} a 28-year-old chef who \textbf{\st{had}} recently moved to San Francisco was found dead \textbf{on} the \textbf{steps} of a local \textbf{department store}.\\
\midrule
REFERENCE & But the victim's brother says he can't think of anyone who would want to hurt him, saying, "Things were finally going well for him."\\
THIS WORK & But the victim's brother says he can't think of anyone who would want to hurt him, saying, "Things were finally going well for him."\\
PARABANK 2 & But the victim's brother said he \textbf{couldn't} think of anyone who\textbf{'d} want to hurt him, \textbf{and he said he was finally okay.}\\
\midrule
REFERENCE & The body found at the Westfield Mall Wednesday morning was identified as 28-year-old San Francisco resident Frank Galicia, the San Francisco Medical Examiner's Office said.\\
THIS WORK & The body found at \textbf{\st{the}} Westfield Mall Wednesday morning was identified as 28-year-old San Francisco resident Frank Galicia, the San Francisco Medical Examiner's Office said.\\
PARABANK 2 &   The body found \textbf{Wednesday morning} at the Westfield Mall \textbf{has been} identified \textbf{by the San Francisco Medical Examiner's Office} as 28-year-old \textcolor{red}{\textbf{\st{San Franscisco resident}}} Frank Galicia.\\
\midrule
REFERENCE & The San Francisco Police Department said the death was ruled a homicide and an investigation is ongoing.\\
THIS WORK & The San Francisco Police Department said the death was \textbf{deemed} a homicide and an investigation is ongoing.\\
PARABANK 2 & \textbf{\st{The}} San Francisco \textbf{P.D. says} the death \textbf{has been ruled} a \textbf{murder} and \textbf{is under investigation}.\\
\midrule
REFERENCE & The victim's brother, Louis Galicia, told ABC station KGO in San Francisco that Frank, previously a line cook in Boston, had landed his dream job as line chef at San Francisco's Sons \& Daughters restaurant six months ago.\\
THIS WORK & The victim's brother, Louis Galicia, told ABC station KGO in San Francisco that Frank, formerly a line cook in Boston, had landed his dream job as line chef at San Francisco's Sons \& Daughters restaurant six months ago.\\
PARABANK 2 & \textbf{\st{The}} Victim's brother, Louis Galicia, told ABC station KGO in San Francisco that Frank, \textbf{who used to be} a line chef in Boston, \textcolor{red}{\textbf{quit}} his dream job \textbf{six months ago} as a line chef at the Sons \& Daughters Restaurant in San Francisco.\\
\midrule
REFERENCE & A spokesperson for Sons \& Daughters said they were "shocked and devastated" by his death.\\
THIS WORK & A spokesperson for Sons \& Daughters said they were "shocked and devastated" by his death\\
PARABANK 2 & A \textbf{spokesman} for Sons \& Daughters said that \textbf{his death} "shocked and devastated \textbf{them.}"\\
\midrule
REFERENCE & "We are a small team that operates like a close knit family and he will be dearly missed," the spokesperson said.\\
THIS WORK & "We are a small team that operates like a close-knit family and he will be dearly missed," the \textbf{spokesman} said.\\
PARABANK 2 & "We are a small team, \textbf{operating} as a close-knit family\textbf{,} and \textbf{we will miss him} dearly," said the \textbf{spokesman}.\\
\midrule
REFERENCE & Our thoughts and condolences are with Frank's family and friends at this difficult time.\\
THIS WORK &   Our thoughts and condolences are with Frank's family and friends at this difficult time.\\
PARABANK 2 &   Our thoughts and condolences go out to Frank's family and friends in these difficult times.\\
\midrule
REFERENCE & Louis Galicia said Frank initially stayed in hostels, but recently, "Things were finally going well for him."\\
THIS WORK & Louis Galicia said Frank initially stayed in hostels, but recently, "Things were finally going well for him."\\
PARABANK 2 & Louis Galicia said \textbf{that} Frank initially stayed in \textcolor{red}{\textbf{the dormitory}}, but \textbf{lately}, "\textbf{He's finally doing okay}."\\
\bottomrule
\end{tabular}%
\caption{Sentences generated via beam search (beamwidth 5) for the multilingual model presented in this work vs ParaBank 2.
  We note that our model tends to produces copies or near copies of the input, which is the desired behavior for our application.
Changes are emphasized with \textbf{bold} or \textbf{\st{strikethrough}}.
 The model trained on ParaBank 2 tends to produce output with lexical/syntactic changes, 
which occasionally also significantly change the meaning of the sentence
(denoted in \textcolor{red}{red}).
 References (paraphraser inputs) are the first ten sentences of WMT17 zh--en.}\label{tab:genexamples}
\end{figure*}

\begin{table}[ht]
\centering
\footnotesize
\begin{tabular}{l r r r}
\toprule

                    & ParaBank 2  & This Work  \\
\midrule
$H(BS|r0)$          &  -0.501    &  -0.225    \\
$H(r0|r0)$          &  -1.157    &  -0.303    \\
$H(r1|r0)$          &  -2.246    &  -2.187    \\
\midrule
$BLEU(BS,r0)$       &  31.9      &  82.8      \\
\bottomrule
\end{tabular}%
\caption[Paraphraser Bias]{
  Average token log probability ($H$) for a sequence generated via beam search ($BS$), 
  a copy of the input ($r0$), and a high-quality human paraphrase of the input ($r1$),
  for a generative paraphraser vs our model, conditioned on $r0$ in all cases.
  BLEU is also computed for the beam search output of each model, with respect to $r0$. Note that BLEU for $r1$ with respect to $r0$ is 17.1.
}\label{fig:paraphraser_input-bias}
\end{table}

\clearpage
\section{Data Details for Replication}\label{appendix:datadetails}
Much of our data comes from WikiMatrix
\cite{DBLP:journals/corr/abs-1907-05791},
a large collection of parallel data extracted from Wikipedia,
and for more domain variety, we added
Global Voices,%
\footnote{\url{http://casmacat.eu/corpus/global-voices.html}}
EuroParl \cite{koehn2005europarl}
(random subset of to 100k sentence pairs per language pair),
SETimes,\footnote{\url{http://nlp.ffzg.hr/resources/corpora/setimes/}}
United Nations \cite{eisele2010multiun}
(random sample of 1M sentence pairs per language pair).
We also included WMT Kazakh--English and Kazakh--Russian data from WMT,
to be able to evaluate on Kazakh.

WMT Kazakh--English and Kazakh--Russian
were limited to the best 1M and 200k sentence pairs, respectively,
as judged by LASER. 
We used a margin threshold of 1.05 for WikiMatrix
and a threshold of 1.04 for the remaining datasets, as we expect them to be cleaner.
We find that FastText classifies many sentences as non-English
  when they contain mostly English but also contain a few non-English words, especially from lower resource languages.
To remedy this, we performed language identification (LID) on 5-grams and filtered out sentences
for which LID did not classify at least half of the 5-grams
as the expected language.

We filtered out sentences where there was more than
60\% overlap in 3-grams
or 40\% overlap in 4-grams.
Via manual inspection, this seemed to provide a good trade-off
between allowing numbers and named entities to be copied,
and filtering out sentences that were clearly not translated.
We perform tokenization with SentencePiece \cite{kudo-richardson-2018-sentencepiece} prior to filtering,
using a 200k vocabulary for all language pairs,
to account for languages like Chinese which do not denote word boundaries.
Note that this vocabulary was used only for filtering,
not for training the final model.

\begin{figure*}[ht!]
\scriptsize
\begin{tikzpicture}
\begin{axis}[
    symbolic x coords={en, es, fr, ru, pt, de, it, ar, zh, cs, el, ro, bg, nl, pl, ca, uk, sv, hu, fi, da, mk, sk, et, tr, lt, vi, sl, id, ja, sq, lv, no, sr, he, kk, eo, hr, bn},
    xtick=data,
    width=\linewidth,
    enlarge x limits=0.01,
    height=8cm,
    scaled y ticks=false,
    ytick={0, 5000000, 10000000, 15000000, 20000000, 25000000, 30000000},
    ymajorgrids,
    yticklabels={0, 2.5M, 5M, 7.5M, 10M, 12.5M, 15M},
    ]
    \addplot[ybar,fill=blue,
    ] coordinates {
(en,33257902)
(es,17349390)
(fr,14758123)
(ru,12875654)
(pt,7963328)
(de,7828466)
(it,7671311)
(ar,7481367)
(zh,6301002)
(cs,5451572)
(el,5184553)
(ro,5131348)
(bg,4650901)
(nl,4475299)
(pl,4135694)
(ca,4000402)
(uk,3781528)
(sv,3623479)
(hu,3397574)
(fi,3075831)
(da,2877977)
(mk,2714589)
(sk,2694277)
(et,2561873)
(tr,2475395)
(lt,2458850)
(vi,2189316)
(sl,2188182)
(id,2185274)
(ja,1928041)
(sq,1894265)
(lv,1800571)
(no,1763623)
(sr,1538184)
(he,1342283)
(kk,1306081)
(eo,1185593)
(hr,1082136)
(bn,1058430)
    };
\end{axis}
\end{tikzpicture}\caption[Paraphraser Training Data Statistics]{Distribution of the 39 languages (ISO~639-1 language code) of the 99.8M training sentences. English accounts for 16.7\%. Spanish, French, Russian, Portuguese, German, and Italian account for a combined 34.3\%. The bottom 20 languages account for only 21.9\% combined.}\label{fig:paraphraser_counts}
\end{figure*}
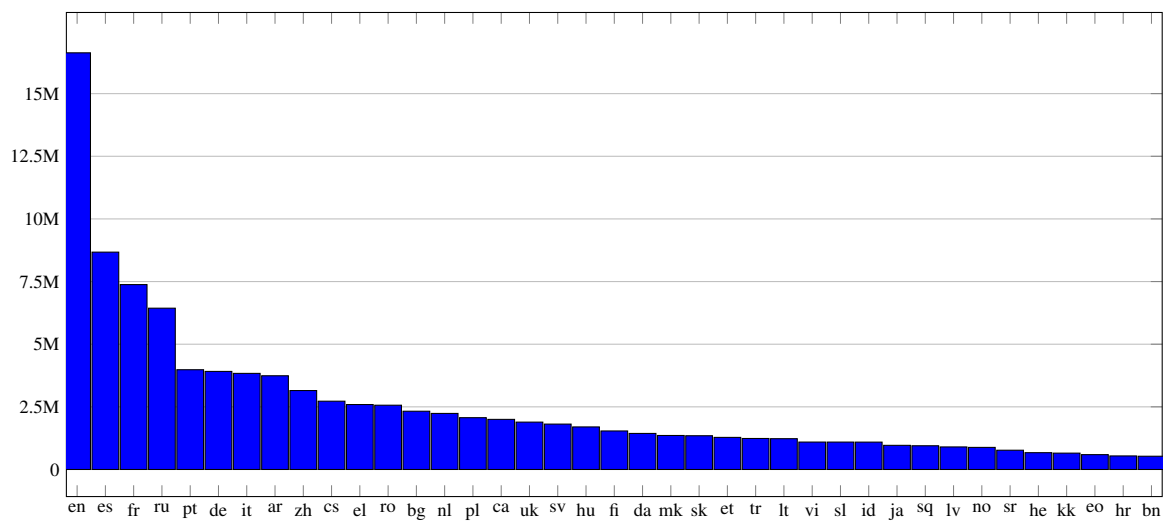

We limited training to languages with at least 1M examples,
which resulted in 39 languages. \autoref{fig:paraphraser_counts} shows the languages and amount of data in each language.

\clearpage
\section{Model Training Details for Replication}\label{appendix:training}

\subsection{Primary Model}

We train a SentencePiece \cite{kudo-richardson-2018-sentencepiece}
model with a 64k vocabulary size on the concatenation of all data,
and filter sentences with length greater than 200 subwords.
Multilingual NMT performance
has been found to increase significantly with model size -- tor example, the best performance of \citet{NIPS2019_8305} is with their largest model which has 6 billion parameters.
Training such a model is well beyond the scope of this work, but
we train a model as large a feasible given our compute budget constraints.
We train a Transformer \cite{vaswani2017attention}
in fairseq \cite{ott2019fairseq} with eight encoder layers,
eight decoder layers,
an embedding size of 1280,
feed forward layer size of 12288,
20 attention heads,
learning rate of $0.0004$,
batch size of 1800 tokens with gradient accumulation over 200 batches,
gradient clipping of 1.2,
and dropout of 0.1. 
The model has approximately 745M parameters for 39 languages.
We train for 6 epochs,
which takes approximately 9 days
on a \verb|p3.16xlarge| instance rented from Amazon AWS,
which has 8 Volta V100 GPUs with 16 GB of memory each. 
No hyperparameters were swept, as training a single model used the majority of our compute budget (the total cost for training this model was approximately \$13,000 USD).
However, we did restart training after discovering that LID was not performing well and adding the 5-gram LID filtering.

\subsection{ParaBank 2 Model}

We train a contrastive, English-only paraphraser on the ParaBank 2 dataset 
\cite{hu-etal-2019-large}. 
We train a Transformer with an 8-layer encoder, 8-layer decoder,
$1024$ dimensional embeddings, embedding sizes of $1024$,
feed-forward size of $4096$,
and $16$ attention heads.
We use a SentencePiece 
model with a 16k vocabulary size.
Dropout is $0.3$, label smoothing is $0.1$, and learning rate is $0.0005$.
The model has approximately 253M parameters for 1 language. 
Batch size is 31200 tokens, and the model trains for approximately 6 weeks (33 epochs) on 4 Nvidia 2080 GPUs.

\subsection{Language Model}

We train a multilingual language model
on the same data as our multilingual NMT system.

The model architecture is based on GPT-2 \cite{radford2019language}, and we use
the fairseq \verb|transformer_lm_gpt2_small| implementation.
We train for 200k updates (18 epochs) of approximately 131k tokens. 
The model has 369M parameters for 39 languages. %
We train with shared embeddings and a learning rate of $0.0005$,
and we stop gradients at sentence boundaries,
using \verb|--sample-break-mode eos|
as the model will be used to evaluate individual sentences.
Other parameters match the fairseq defaults.
The model trained for approximately 4 weeks
on 4 Nvidia TITAN RTX GPUs.

\subsection{Autoencoder}

We use the pretrained ``multilingual denoising pre-trained model'' (mBART) model 
of \citet{liu-etal-2020-multilingual}, as it works in all languages of interest.
Their model is designed to be fine-tuned to translation tasks, and their fine-tuning introduces subtle changes to the decoder that are required for inference.
In order to adapt it to our task, we therefore fine-tune for a single update with a learning rate of 0.
We then produce scores with the model in the same manner as Prism-ref.
The model has approximately 680M parameters for 25 languages.
We did not train this model but note that doing so required substantial compute power -- \citet{liu-etal-2020-multilingual}
note that they trained for approximately 2.5 weeks on 256 Nvidia V100 GPUS, each with 32GB of memory. 

\subsection{Baselines}

We compare to BLEURT \cite{bleurt} using the authors' recommended ``BLEURT-Base 128''\footnote{\url{https://github.com/google-research/bleurt}}
We compare to BERTscore F1 \cite{bert-score} using the model and code provided by the authors.\footnote{\url{https://github.com/Tiiiger/bert_score}}
The remaining baseline results are computed using the metric scores as submitted to \cite{ma-etal-2019-results}\footnote{\url{http://data.statmt.org/wmt19/translation-task/wmt19-submitted-data-v3.tgz}}

\clearpage
\section{WMT 2018 (Development set) Results: System-level, Segment-level, and Sweeps}\label{appendix:prelimresults}

\begin{figure}[h!]
\centering
\includegraphics[width=0.7\linewidth]{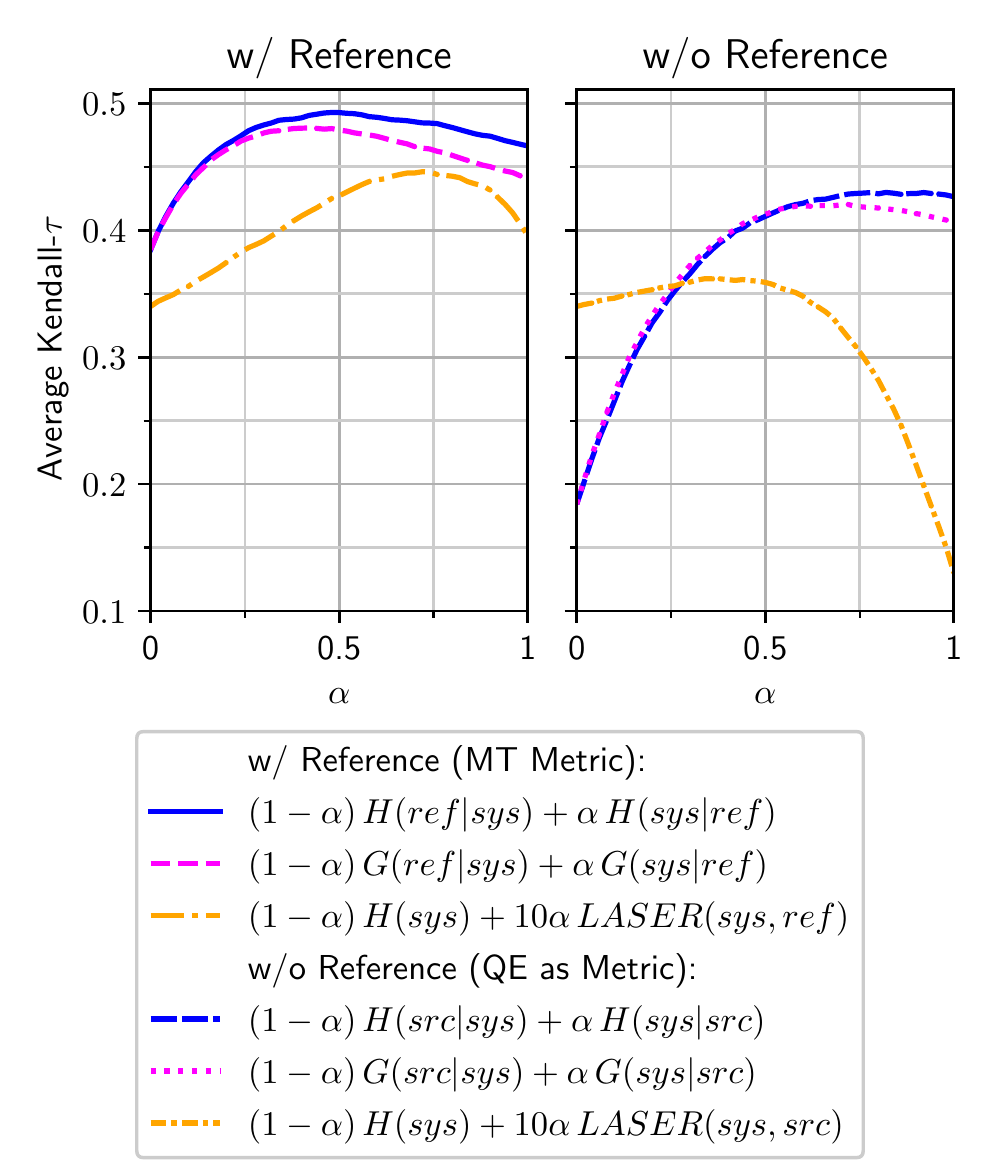}
\caption{Linear combinations of scoring each direction
  using length-normalized ($H$) vs un-normalized ($G$) log probability
  for our method,
  and length-normalized language model probabilities ($H$) vs
  LASER for our contrastive method.
  In both cases, we explore scoring using the human reference $\reff$
  vs the source $\src$. Results are segment-level $\tau$
  on our development set (WMT18), averaged across all language pairs.
}%
\label{fig:results:prelim}
\end{figure}

\autoref{fig:results:prelim} shows results on the development set (WMT18) for sweeping various linear combinations.

\autoref{fig:mt_metric_wmt18_Segment_toEnglish},
\autoref{fig:mt_metric_wmt18_Segment_fromEnglish},
\autoref{fig:mt_metric_wmt18_System_toEnglish}
and
\autoref{fig:mt_metric_wmt18_System_fromEnglish},
show full segment- and system- level results, into and out of English,
for the WMT 2018 MT metrics shared task, along with
all baselines and submitted systems.

\begin{table*}[h!]
\centering
\footnotesize
    \addtolength{\tabcolsep}{-2pt} 
\begin{tabular}{lrrrrrrrrrrrrrrrrrrrrrrrrrrrrrrrrrrrrrrrrrrrrrrrrr}
\toprule
                                                                                          & {\bf cs--en}  & {\bf de--en} & {\bf et--en} & {\bf fi--en} & {\bf ru--en} & {\bf tr--en}  & {\bf zh--en} \\
n                                                                                         &     5110     &    77811    &    56721    &    15648    &    10404    &     8525     &    33357    \\
\midrule
\nonen \metric{BEER}${}^\submission$               \cite{stanojevic-simaan-2015-beer}     &       0.295  &      0.481  &      0.341  &      0.232  &      0.288  &      0.229   &      0.214  \\
\nonen \metric{BERTscore}      \cite{bertscore_arxiv,bert-score}                          &  {\bf 0.404} &      0.550  & {\bf 0.397} & {\bf 0.296} & {\bf 0.340} & {\bf 0.292}  & {\bf 0.253} \\
\ensem \metric{BLEND}${}^\submission$              \cite{ma-etal-2017-blend}              &       0.322  &      0.492  &      0.354  &      0.226  &      0.290  &      0.232   &      0.217  \\
\nonen \metric{CharacTER}${}^\submission$          \cite{wang-etal-2016-character}        &       0.256  &      0.450  &      0.286  &      0.185  &      0.244  &      0.172   &      0.202  \\
\nonen \metric{chrF}${}^\baseline$               \cite{popovic-2015-chrf}                 &       0.288  &      0.479  &      0.328  &      0.229  &      0.269  &      0.210   &      0.208  \\
\nonen \metric{chrF+}${}^\baseline$              \cite{popovic-2017-chrf}                 &       0.288  &      0.479  &      0.332  &      0.234  &      0.279  &      0.218   &      0.207  \\
\nonen \metric{ITER}${}^\submission$               \cite{panja-naskar-2018-iter}          &       0.198  &      0.396  &      0.235  &      0.128  &      0.139  &      -0.029  &      0.144  \\
\nonen \metric{meteor++}${}^\submission$           \cite{shimanaka-etal-2018-ruse}        &       0.270  &      0.457  &      0.329  &      0.207  &      0.253  &      0.204   &      0.179  \\
\nonen \metric{RUSE}${}^\submission$               \cite{shimanaka-etal-2018-ruse}        &       0.347  &      0.498  &      0.368  &      0.273  &      0.311  &      0.259   &      0.218  \\
\nonen \metric{sentBLEU}${}^\baseline$           \cite{papineni-etal-2002-bleu}           &       0.233  &      0.415  &      0.285  &      0.154  &      0.228  &      0.145   &      0.178  \\
\nonen \metric{UHH\_TSKM}${}^\submission$          \cite{duma-menzel-2017-uhh-submission} &       0.274  &      0.436  &      0.300  &      0.168  &      0.235  &      0.154   &      0.151  \\
\nonen \metric{YiSi-0}${}^\submission$             \cite{lo-2019-yisi}                    &       0.301  &      0.474  &      0.330  &      0.225  &      0.294  &      0.215   &      0.205  \\
\nonen \metric{YiSi-1}${}^\submission$             \cite{lo-2019-yisi}                    &       0.319  &      0.488  &      0.351  &      0.231  &      0.300  &      0.234   &      0.211  \\
\nonen \metric{YiSi-1\_srl}${}^\submission$        \cite{lo-2019-yisi}                    &       0.317  &      0.483  &      0.345  &      0.237  &      0.306  &      0.233   &      0.209  \\
\midrule
\nonen \BJTwikiplussysref                                                                 & {\bf 0.423}  & {\bf 0.560} & {\bf 0.409} & {\bf 0.317} & {\bf 0.366} & {\bf 0.309}  & {\bf 0.263} \\
\nonen \BJTParabanksysref                                                                 & {\bf 0.386}  &      0.538  & {\bf 0.399} & {\bf 0.309} & {\bf 0.340} &      0.275   &      0.244  \\
\nonen \BJTLASERrefLM                                                                     &      0.364   &      0.526  &      0.378  &      0.265  &      0.305  &      0.257   &      0.243  \\
\nonen \BJTwikiplussysgivensrc                                                            &      0.355   &      0.515  &      0.370  &      0.257  &      0.308  &      0.213   &      0.194  \\
\nonen \BJTwikipluslm                                                                     &      0.285   &      0.438  &      0.285  &      0.198  &      0.280  &      0.123   &      0.192  \\
\nonen \BJTLASERref                                                                       &      0.310   &      0.494  &      0.364  &      0.232  &      0.257  &      0.248   &      0.207  \\
\nonen \BJTmBARTsysref                                                                    &      0.251   &      0.455  &      0.315  &      0.199  &      0.248  &      0.196   &      0.181  \\
\bottomrule
\end{tabular}\caption[WMT18 Segment-level Results, to English]{WMT18 Segment-level results, to English. n denotes number of pairwise judgments. \BoldDenotes{ }We exclude BLEURT \cite{bleurt} as it was directly trained on WMT18 judgements. $\baseline$:WMT18 Baseline \cite{ma-etal-2018-results} $\submission$:WMT18 Metric Submission \cite{ma-etal-2018-results}}\label{fig:mt_metric_wmt18_Segment_toEnglish}
\end{table*}

\begin{table*}[h!]
\centering
  \footnotesize
   \addtolength{\tabcolsep}{-2pt} 
\begin{tabular}{lrrrrrrrrrrrrrrrrrrrrrrrrrrrrrrrrrrrrrrrrrrrrrrrrr}
\toprule
                                                                                      & {\bf en--cs}  & {\bf en--de} & {\bf en--et} & {\bf en--fi} & {\bf en--ru} & {\bf en--tr} & {\bf en--zh} \\
n                                                                                     &     5413     &    19711    &    32202    &     9809    &    22181    &     1358    &    28602    \\
\midrule
\nonen \metric{BEER}${}^\submission$               \cite{stanojevic-simaan-2015-beer} &       0.518  &      0.686  &      0.558  &      0.511  &      0.403  &      0.374  &      0.302  \\
\nonen \metric{BERTscore}      \cite{bertscore_arxiv,bert-score}                      &       0.559  &      0.727  &      0.584  &      0.538  &      0.424  &      0.389  & {\bf 0.364} \\
\ensem \metric{BLEND}${}^\submission$              \cite{ma-etal-2017-blend}          &          $-$ &         $-$ &         $-$ &         $-$ &      0.394  &         $-$ &         $-$ \\
\nonen \metric{CharacTER}${}^\submission$          \cite{wang-etal-2016-character}    &       0.414  &      0.604  &      0.464  &      0.403  &      0.352  &      0.404  &      0.313  \\
\nonen \metric{chrF}${}^\baseline$               \cite{popovic-2015-chrf}             &       0.516  &      0.677  &      0.572  &      0.520  &      0.383  &      0.409  &      0.328  \\
\nonen \metric{chrF+}${}^\baseline$              \cite{popovic-2017-chrf}             &       0.513  &      0.680  &      0.573  &      0.525  &      0.392  &      0.405  &      0.328  \\
\nonen \metric{ITER}${}^\submission$               \cite{panja-naskar-2018-iter}      &       0.333  &      0.610  &      0.392  &      0.311  &      0.291  &      0.236  &         $-$ \\
\nonen \metric{sentBLEU}${}^\baseline$           \cite{papineni-etal-2002-bleu}       &       0.389  &      0.620  &      0.414  &      0.355  &      0.330  &      0.261  &      0.311  \\
\nonen \metric{YiSi-0}${}^\submission$             \cite{lo-2019-yisi}                &       0.471  &      0.661  &      0.531  &      0.464  &      0.394  &      0.376  &      0.318  \\
\nonen \metric{YiSi-1}${}^\submission$             \cite{lo-2019-yisi}                &       0.496  &      0.691  &      0.546  &      0.504  &      0.407  &      0.418  &      0.323  \\
\nonen \metric{YiSi-1\_srl}${}^\submission$        \cite{lo-2019-yisi}                &          $-$ &      0.696  &         $-$ &         $-$ &         $-$ &         $-$ &      0.310  \\
\midrule
\nonen \BJTwikiplussysref                                                             & {\bf 0.667}  & {\bf 0.799} & {\bf 0.705} & {\bf 0.667} & {\bf 0.469} & {\bf 0.574} & {\bf 0.371} \\
\nonen \BJTLASERrefLM                                                                 &      0.587   &      0.746  &      0.628  &      0.629  &      0.450  & {\bf 0.501} & {\bf 0.367} \\
\nonen \BJTwikiplussysgivensrc                                                        &      0.552   &      0.732  &      0.636  &      0.626  &      0.409  & {\bf 0.505} &      0.298  \\
\nonen \BJTwikipluslm                                                                 &      0.459   &      0.655  &      0.408  &      0.511  &      0.375  &      0.331  &      0.221  \\
\nonen \BJTLASERref                                                                   &      0.480   &      0.677  &      0.585  &      0.511  &      0.402  &      0.432  &      0.338  \\
\nonen \BJTmBARTsysref                                                                &      0.404   &      0.594  &      0.405  &      0.410  &      0.356  &      0.303  &      0.305  \\
\bottomrule
\end{tabular}\caption[WMT18 Segment-level Results, from English]{WMT18 Segment-level results, from English. n denotes number of pairwise judgments. \BoldDenotes{ }$\baseline$:WMT18 Baseline \cite{ma-etal-2018-results} $\submission$:WMT18 Metric Submission \cite{ma-etal-2018-results}}\label{fig:mt_metric_wmt18_Segment_fromEnglish}
\end{table*}

\begin{table*}[h!]
\centering
\footnotesize
    \addtolength{\tabcolsep}{-2pt}
\begin{tabular}{lrrrrrrrrrrrrrrrrrrrrrrrrrrrrrrrrrrrrrrrrrrrrrrrrr}
\toprule
                                                                                          & {\bf cs--en}  & {\bf de--en} & {\bf et--en} & {\bf fi--en} & {\bf ru--en} & {\bf tr--en} & {\bf zh--en} \\
n                                                                                         &          5   &         16  &         14  &          9  &          8  &          5  &         14  \\
\midrule
\nonen \metric{BEER}${}^\submission$               \cite{stanojevic-simaan-2015-beer}     &       0.958  &      0.994  & {\bf 0.985} &      0.991  &      0.982  &      0.870  & {\bf 0.976} \\
\nonen \metric{BERTscore}      \cite{bertscore_arxiv,bert-score}                          &  {\bf 0.990} & {\bf 0.999} & {\bf 0.990} & {\bf 0.998} &      0.935  &      0.499  &      0.956  \\
\ensem \metric{BLEND}${}^\submission$              \cite{ma-etal-2017-blend}              &       0.973  &      0.991  &      0.985  & {\bf 0.994} & {\bf 0.993} & {\bf 0.801} & {\bf 0.976} \\
\nonen \metric{BLEU}${}^\baseline$               \cite{papineni-etal-2002-bleu}           &       0.970  &      0.971  & {\bf 0.986} &      0.973  &      0.979  & {\bf 0.657} & {\bf 0.978} \\
\nonen \metric{CDER}${}^\baseline$               \cite{leusch-etal-2006-cder}             &       0.972  &      0.980  & {\bf 0.990} &      0.984  &      0.980  & {\bf 0.664} & {\bf 0.982} \\
\nonen \metric{CharacTER}${}^\submission$          \cite{wang-etal-2016-character}        &  {\bf 0.970} &      0.993  &      0.979  &      0.989  & {\bf 0.991} & {\bf 0.782} &      0.950  \\
\nonen \metric{chrF}${}^\baseline$               \cite{popovic-2015-chrf}                 &       0.966  &      0.994  &      0.981  &      0.987  & {\bf 0.990} &      0.452  &      0.960  \\
\nonen \metric{chrF+}${}^\baseline$              \cite{popovic-2017-chrf}                 &       0.966  &      0.993  &      0.981  &      0.989  &      0.990  &      0.174  &      0.964  \\
\nonen \metric{ITER}${}^\submission$               \cite{panja-naskar-2018-iter}          &       0.975  &      0.990  &      0.975  & {\bf 0.996} &      0.937  & {\bf 0.861} & {\bf 0.980} \\
\nonen \metric{meteor++}${}^\submission$           \cite{shimanaka-etal-2018-ruse}        &       0.945  &      0.991  &      0.978  &      0.971  & {\bf 0.995} &      0.864  &      0.962  \\
\nonen \metric{NIST}${}^\baseline$               \cite{doddington2002automatic}           &       0.954  &      0.984  &      0.983  &      0.975  &      0.973  & {\bf 0.970} &      0.968  \\
\nonen \metric{PER}${}^\baseline$                                                         & {\bf 0.970}  &      0.985  & {\bf 0.983} &      0.993  &      0.967  &      0.159  &      0.931  \\
\nonen \metric{RUSE}${}^\submission$               \cite{shimanaka-etal-2018-ruse}        &       0.981  &      0.997  & {\bf 0.990} &      0.991  & {\bf 0.988} & {\bf 0.853} & {\bf 0.981} \\
\nonen \metric{TER}${}^\baseline$                \cite{snover2006study}                   &  {\bf 0.950} &      0.970  & {\bf 0.990} &      0.968  &      0.970  &      0.533  &      0.975  \\
\nonen \metric{UHH\_TSKM}${}^\submission$          \cite{duma-menzel-2017-uhh-submission} &       0.952  &      0.980  & {\bf 0.989} &      0.982  &      0.980  &      0.547  & {\bf 0.981} \\
\nonen \metric{WER}${}^\baseline$                                                         & {\bf 0.951}  &      0.961  & {\bf 0.991} &      0.961  &      0.968  &      0.041  &      0.975  \\
\nonen \metric{YiSi-0}${}^\submission$             \cite{lo-2019-yisi}                    &       0.956  &      0.994  &      0.975  &      0.978  &      0.988  & {\bf 0.954} &      0.957  \\
\nonen \metric{YiSi-1}${}^\submission$             \cite{lo-2019-yisi}                    &       0.950  &      0.992  &      0.979  &      0.973  & {\bf 0.991} & {\bf 0.958} &      0.951  \\
\nonen \metric{YiSi-1\_srl}${}^\submission$        \cite{lo-2019-yisi}                    &       0.965  &      0.995  &      0.981  &      0.977  & {\bf 0.992} & {\bf 0.869} &      0.962  \\
\midrule
\nonen \BJTwikiplussysref                                                                 &      0.988   &      0.995  &      0.971  & {\bf 0.998} & {\bf 0.995} &      0.730  & {\bf 0.989} \\
\nonen \BJTParabanksysref                                                                 & {\bf 0.992}  &      0.989  &      0.964  & {\bf 0.998} & {\bf 0.996} & {\bf 0.896} &      0.986  \\
\nonen \BJTLASERrefLM                                                                     & {\bf 0.988}  &      0.991  &      0.965  &      0.994  &      0.745  &      0.297  &      0.890  \\
\nonen \BJTwikiplussysgivensrc                                                            & {\bf 0.984}  &      0.991  &      0.964  &      0.987  &      0.970  & {\bf 0.896} &      0.958  \\
\nonen \BJTwikipluslm                                                                     & {\bf 0.986}  &      0.970  &      0.954  &      0.898  &      0.951  & {\bf 0.891} &      0.972  \\
\nonen \BJTLASERref                                                                       & {\bf 0.978}  &      0.986  &      0.953  &      0.984  &      0.489  & {\bf 0.968} &      0.591  \\
\nonen \BJTmBARTsysref                                                                    &      0.955   &      0.996  & {\bf 0.987} &      0.995  &      0.981  &      0.721  & {\bf 0.980} \\
\bottomrule
\end{tabular}\caption[WMT18 System-level Results, to English]{WMT18 System-level results, to English. n denotes number of MT systems. \BoldDenotes{ }We exclude BLEURT \cite{bleurt} as it was directly trained on WMT18 judgements. $\baseline$:WMT18 Baseline \cite{ma-etal-2018-results} $\submission$:WMT18 Metric Submission \cite{ma-etal-2018-results}}\label{fig:mt_metric_wmt18_System_toEnglish}
\end{table*}

\begin{table*}[h!]
\centering
\footnotesize
    \addtolength{\tabcolsep}{-2pt} 
\begin{tabular}{lrrrrrrrrrrrrrrrrrrrrrrrrrrrrrrrrrrrrrrrrrrrrrrrrr}
\toprule
                                                                                      & {\bf en--cs}  & {\bf en--de} & {\bf en--et} & {\bf en--fi} & {\bf en--ru} & {\bf en--tr} & {\bf en--zh} \\
n                                                                                     &          5   &         16  &         14  &         12  &          9  &          8  &         14  \\
\midrule
\nonen \metric{BEER}${}^\submission$               \cite{stanojevic-simaan-2015-beer} &  {\bf 0.992} & {\bf 0.991} & {\bf 0.980} & {\bf 0.961} & {\bf 0.988} & {\bf 0.965} &      0.928  \\
\nonen \metric{BERTscore}      \cite{bertscore_arxiv,bert-score}                      &  {\bf 0.997} & {\bf 0.989} & {\bf 0.982} & {\bf 0.972} & {\bf 0.990} &      0.908  &      0.967  \\
\ensem \metric{BLEND}${}^\submission$              \cite{ma-etal-2017-blend}          &          $-$ &         $-$ &         $-$ &         $-$ & {\bf 0.988} &         $-$ &         $-$ \\
\nonen \metric{BLEU}${}^\baseline$               \cite{papineni-etal-2002-bleu}       &       0.995  &      0.981  &      0.975  & {\bf 0.962} &      0.983  &      0.826  &      0.947  \\
\nonen \metric{CDER}${}^\baseline$               \cite{leusch-etal-2006-cder}         &       0.997  &      0.986  & {\bf 0.984} & {\bf 0.964} & {\bf 0.984} &      0.861  &      0.961  \\
\nonen \metric{CharacTER}${}^\submission$          \cite{wang-etal-2016-character}    &  {\bf 0.993} & {\bf 0.989} &      0.956  & {\bf 0.974} &      0.983  &      0.833  & {\bf 0.983} \\
\nonen \metric{chrF}${}^\baseline$               \cite{popovic-2015-chrf}             &  {\bf 0.990} & {\bf 0.990} & {\bf 0.981} & {\bf 0.969} & {\bf 0.989} &      0.948  &      0.944  \\
\nonen \metric{chrF+}${}^\baseline$              \cite{popovic-2017-chrf}             &  {\bf 0.990} &      0.989  & {\bf 0.982} & {\bf 0.970} &      0.989  &      0.943  &      0.943  \\
\nonen \metric{ITER}${}^\submission$               \cite{panja-naskar-2018-iter}      &       0.915  & {\bf 0.984} & {\bf 0.981} & {\bf 0.973} &      0.975  &      0.865  &         $-$ \\
\nonen \metric{NIST}${}^\baseline$               \cite{doddington2002automatic}       &  {\bf 0.999} &      0.986  & {\bf 0.983} &      0.949  & {\bf 0.990} &      0.902  &      0.950  \\
\nonen \metric{PER}${}^\baseline$                                                     &      0.991   &      0.981  &      0.958  &      0.906  & {\bf 0.988} &      0.859  &      0.964  \\
\nonen \metric{TER}${}^\baseline$                \cite{snover2006study}               &  {\bf 0.997} & {\bf 0.988} & {\bf 0.981} &      0.942  & {\bf 0.987} &      0.867  & {\bf 0.963} \\
\nonen \metric{WER}${}^\baseline$                                                     & {\bf 0.997}  & {\bf 0.986} & {\bf 0.981} &      0.945  &      0.985  &      0.853  &      0.957  \\
\nonen \metric{YiSi-0}${}^\submission$             \cite{lo-2019-yisi}                &       0.973  &      0.985  &      0.968  &      0.944  & {\bf 0.990} & {\bf 0.990} &      0.957  \\
\nonen \metric{YiSi-1}${}^\submission$             \cite{lo-2019-yisi}                &  {\bf 0.987} &      0.985  & {\bf 0.979} &      0.940  & {\bf 0.992} & {\bf 0.976} &      0.963  \\
\nonen \metric{YiSi-1\_srl}${}^\submission$        \cite{lo-2019-yisi}                &          $-$ & {\bf 0.990} &         $-$ &         $-$ &         $-$ &         $-$ &      0.952  \\
\midrule
\nonen \BJTwikiplussysref                                                             &      0.962   & {\bf 0.987} & {\bf 0.973} & {\bf 0.976} & {\bf 0.989} &      0.894  & {\bf 0.977} \\
\nonen \BJTLASERrefLM                                                                 &      0.953   & {\bf 0.984} & {\bf 0.980} & {\bf 0.976} &      0.984  &      0.927  & {\bf 0.982} \\
\nonen \BJTwikiplussysgivensrc                                                        &      0.850   & {\bf 0.984} &      0.949  &      0.964  &      0.960  &      0.864  &      0.940  \\
\nonen \BJTwikipluslm                                                                 &      0.854   & {\bf 0.985} &      0.837  &      0.938  &      0.959  &      0.830  &      0.859  \\
\nonen \BJTLASERref                                                                   & {\bf 0.995}  &      0.965  &      0.937  & {\bf 0.978} & {\bf 0.993} &      0.895  & {\bf 0.978} \\
\nonen \BJTmBARTsysref                                                                &      0.985   & {\bf 0.989} & {\bf 0.977} & {\bf 0.959} & {\bf 0.987} & {\bf 0.963} &      0.689  \\
\bottomrule
\end{tabular}\caption[WMT18 System-level Results, from English]{WMT18 System-level results, from English. n denotes number of MT systems. \BoldDenotes{ }$\baseline$:WMT18 Baseline \cite{ma-etal-2018-results} $\submission$:WMT18 Metric Submission \cite{ma-etal-2018-results}}\label{fig:mt_metric_wmt18_System_fromEnglish}
\end{table*}

\clearpage
\section{WMT 2019 Metric and QE as Metric Segment-Level Results}\label{appendix:wmt19_seg}

\autoref{fig:mt_metric_wmt19_Segment_toEnglish},
\autoref{fig:mt_metric_wmt19_Segment_fromEnglish},
and
\autoref{fig:mt_metric_wmt19_Segment_nonEnglish}
show segment-level metrics (excluding QE as a metric) results, 
for language pairs into, out of, and not including English,
for the WMT 2019 MT metrics shared task, along with
all baselines and submitted systems.

\autoref{fig:mt_metric_wmt19_Segment_toEnglish-qe},
\autoref{fig:mt_metric_wmt19_Segment_fromEnglish-qe},
and
\autoref{fig:mt_metric_wmt19_Segment_nonEnglish-qe}
show segment-level QE as a metric results, 
for language pairs into, out of, and not including English,
for the WMT 2019 MT metrics shared task, along with
all baselines and submitted systems. 

\begin{table*}[h!]
\centering
\footnotesize
    \addtolength{\tabcolsep}{-2pt} 
\begin{tabular}{lrrrrrrrrrrrrrrrrrrrrrrrrrrrrrrrrrrrrrrrrrrrrrrrrr}
\toprule
                                                                                                           & {\bf de--en}     & {\bf fi--en} & {\bf gu--en}    & {\bf kk--en} & {\bf lt--en} & {\bf ru--en}    & {\bf zh--en} \\
n                                                                                                          &      85365      &      38307  &      31139     &      27094  &      21862  &      46172     &      31070  \\
\midrule
\nonen \metric{BEER}${}^\submission$               \cite{stanojevic-simaan-2015-beer}                      &       0.128     &      0.283  &      0.260     &      0.421  &      0.315  &      0.189     &      0.371  \\
\nonen \metric{BERTr}${}^\submission$              \cite{mathur-etal-2019-putting}                         &       0.142     &      0.331  &      0.291     &      0.421  &      0.353  &      0.195     &      0.399  \\
\nonen \metric{BERTscore}      \cite{bertscore_arxiv,bert-score}                                           &       0.176     &       0.345  & {\bf 0.320}    & {\bf 0.432} & {\bf 0.381} & {\bf 0.223}    & {\bf 0.430} \\
\nonen \metric{BLEURT}         \cite{bleurt}                                                               &  {\bf 0.204}    &  {\bf 0.367} & {\bf 0.311}    & {\bf 0.447} & {\bf 0.387} & {\bf 0.228}    & {\bf 0.423} \\
\nonen \metric{CharacTER}${}^\submission$          \cite{wang-etal-2016-character}                         &       0.101     &      0.253  &      0.190     &      0.340  &      0.254  &      0.155     &      0.337  \\
\nonen \metric{chrF}${}^\baseline$               \cite{popovic-2015-chrf}                                  &       0.122     &      0.286  &      0.256     &      0.389  &      0.301  &      0.180     &      0.371  \\
\nonen \metric{chrF+}${}^\baseline$              \cite{popovic-2017-chrf}                                  &       0.125     &      0.289  &      0.257     &      0.394  &      0.303  &      0.182     &      0.374  \\
\nonen \metric{EED}${}^\submission$                \cite{stanchev-etal-2019-eed}                           &       0.120     &      0.281  &      0.264     &      0.392  &      0.298  &      0.176     &      0.376  \\
\nonen \metric{ESIM}${}^\submission$               \cite{chen-etal-2017-enhanced,mathur-etal-2019-putting} &       0.167     &      0.337  &      0.303     & {\bf 0.435} &      0.359  &      0.201     &      0.396  \\
\nonen \metric{hLEPORa\_baseline}${}^\submission$   \cite{han-etal-2012-lepor,han13language}               &          $-$    &         $-$ &         $-$    &      0.372  &         $-$ &         $-$    &      0.339  \\
\nonen \metric{Meteor++\_2.0(syntax)}${}^\submission$   \cite{guo-hu-2019-meteor}                          &       0.084     &      0.274  &      0.237     &      0.395  &      0.291  &      0.156     &      0.370  \\
\nonen \metric{Meteor++\_2.0(syntax+copy)}${}^\submission$   \cite{guo-hu-2019-meteor}                     &       0.094     &      0.273  &      0.244     &      0.402  &      0.287  &      0.163     &      0.367  \\
\nonen \metric{PReP}${}^\submission$               \cite{yoshimura-etal-2019-filtering}                    &       0.030     &      0.197  &      0.192     &      0.386  &      0.193  &      0.124     &      0.267  \\
\nonen \metric{sentBLEU}${}^\baseline$           \cite{papineni-etal-2002-bleu}                            &       0.056     &      0.233  &      0.188     &      0.377  &      0.262  &      0.125     &      0.323  \\
\nonen \metric{WMDO}${}^\submission$               \cite{chow-etal-2019-wmdo}                              &       0.096     &      0.281  &      0.260     &      0.420  &      0.300  &      0.162     &      0.362  \\
\nonen \metric{YiSi-0}${}^\submission$             \cite{lo-2019-yisi}                                     &       0.117     &      0.271  &      0.263     &      0.402  &      0.289  &      0.178     &      0.355  \\
\nonen \metric{YiSi-1}${}^\submission$             \cite{lo-2019-yisi}                                     &       0.164     &      0.347  & {\bf 0.312}    & {\bf 0.440} & {\bf 0.376} & {\bf 0.217}    & {\bf 0.426} \\
\nonen \metric{YiSi-1\_srl}${}^\submission$        \cite{lo-2019-yisi}                                     &  {\bf 0.199}    &      0.346  &      0.306     & {\bf 0.442} & {\bf 0.380} & {\bf 0.222}    & {\bf 0.431} \\
\midrule
\nonen \BJTwikiplussysref                                                                                  & {\bf 0.204}     & {\bf 0.357} & {\bf 0.313}    & {\bf 0.434} & {\bf 0.382} & {\bf 0.225}    & {\bf 0.438} \\
\nonen \BJTParabanksysref                                                                                  &      0.184      & {\bf 0.341} & {\bf 0.326}    &      0.425  & {\bf 0.373} &      0.207     & {\bf 0.432} \\
\nonen \BJTLASERrefLM                                                                                      &      0.190      &      0.335  & {\bf 0.319}    &      0.428  & {\bf 0.368} &      0.207     &      0.416  \\
\nonen \BJTwikipluslm                                                                                      &      0.083      &      0.253  &      0.165     &      0.120  &      0.281  &      0.130     &      0.210  \\
\nonen \BJTLASERref                                                                                        &      0.151      &      0.301  &      0.305     &      0.420  &      0.325  &      0.193     &      0.397  \\
\nonen \BJTmBARTsysref                                                                                      &      0.136      &      0.255  &      0.246     &      0.377  &      0.298  &      0.162     &      0.349  \\
\bottomrule
\end{tabular}\caption[WMT19 Segment-level Results, to English]{WMT19 Segment-level results, metrics (excludes QE as metric), to English. n denotes number of pairwise judgments. \BoldDenotes{ }$\baseline$:WMT19 Baseline \cite{ma-etal-2019-results} $\submission$:WMT19 Metric Submission \cite{ma-etal-2019-results}
}\label{fig:mt_metric_wmt19_Segment_toEnglish}
\end{table*}

\begin{table*}[h!]
\centering
\footnotesize
    \addtolength{\tabcolsep}{-2pt} 
\begin{tabular}{lrrrrrrrrrrrrrrrrrrrrrrrrrrrrrrrrrrrrrrrrrrrrrrrrr}
\toprule
                                                                                                           & {\bf en--cs}     & {\bf en--de}    & {\bf en--fi}    & {\bf en--gu} & {\bf en--kk} & {\bf en--lt}    & {\bf en--ru}    & {\bf en--zh} \\
n                                                                                                          &      27178      &      99840     &      31820     &      11355  &      18172  &      17401     &      24334     &      18658  \\
\midrule
\nonen \metric{BEER}${}^\submission$               \cite{stanojevic-simaan-2015-beer}                      &       0.443     &      0.316     &      0.514     &      0.537  &      0.516  &      0.441     &      0.542     &      0.232  \\
\nonen \metric{BERTscore}      \cite{bertscore_arxiv,bert-score}                                           &       0.485     &      0.345     &      0.524     & {\bf 0.558} & {\bf 0.533} &      0.463     &      0.580     &      0.347  \\
\nonen \metric{CharacTER}${}^\submission$          \cite{wang-etal-2016-character}                         &       0.349     &      0.264     &      0.404     &      0.500  &      0.351  &      0.311     &      0.432     &      0.094  \\
\nonen \metric{chrF}${}^\baseline$               \cite{popovic-2015-chrf}                                  &       0.455     &      0.326     &      0.514     &      0.534  &      0.479  &      0.446     &      0.539     &      0.301  \\
\nonen \metric{chrF+}${}^\baseline$              \cite{popovic-2017-chrf}                                  &       0.458     &      0.327     &      0.514     &      0.538  &      0.491  &      0.448     &      0.543     &      0.296  \\
\nonen \metric{EED}${}^\submission$                \cite{stanchev-etal-2019-eed}                           &       0.431     &      0.315     &      0.508     & {\bf 0.568} &      0.518  &      0.425     &      0.546     &      0.257  \\
\nonen \metric{ESIM}${}^\submission$               \cite{chen-etal-2017-enhanced,mathur-etal-2019-putting} &          $-$    &      0.329     &      0.511     &         $-$ &      0.510  &      0.428     &      0.572     &      0.339  \\
\nonen \metric{hLEPORa\_baseline}${}^\submission$   \cite{han-etal-2012-lepor,han13language}               &          $-$    &         $-$    &         $-$    &      0.463  &      0.390  &         $-$    &         $-$    &         $-$ \\
\nonen \metric{sentBLEU}${}^\baseline$           \cite{papineni-etal-2002-bleu}                            &       0.367     &      0.248     &      0.396     &      0.465  &      0.392  &      0.334     &      0.469     &      0.270  \\
\nonen \metric{YiSi-0}${}^\submission$             \cite{lo-2019-yisi}                                     &       0.406     &      0.304     &      0.483     &      0.539  &      0.494  &      0.402     &      0.535     &      0.266  \\
\nonen \metric{YiSi-1}${}^\submission$             \cite{lo-2019-yisi}                                     &       0.475     &      0.351     &      0.537     & {\bf 0.551} & {\bf 0.546} &      0.470     &      0.585     & {\bf 0.355} \\
\nonen \metric{YiSi-1\_srl}${}^\submission$        \cite{lo-2019-yisi}                                     &          $-$    &      0.368     &         $-$    &         $-$ &         $-$ &         $-$    &         $-$    & {\bf 0.361} \\
\midrule
\nonen \BJTwikiplussysref                                                                                  & {\bf 0.582}     & {\bf 0.427}    & {\bf 0.591}    &      0.313  & {\bf 0.531} & {\bf 0.558}    &      0.584     & {\bf 0.376} \\
\nonen \BJTLASERrefLM                                                                                      &      0.535      &      0.401     &      0.568     &      0.306  &      0.408  &      0.503     & {\bf 0.640}    & {\bf 0.356} \\
\nonen \BJTwikipluslm                                                                                      &      0.439      &      0.329     &      0.477     &      0.181  &      0.284  &      0.430     &      0.586     &      0.279  \\
\nonen \BJTLASERref                                                                                        &      0.408      &      0.334     &      0.509     &      0.340  &      0.363  &      0.396     &      0.511     &      0.284  \\
\nonen \BJTmBARTsysref                                                                                      &      0.345      &      0.302     &      0.401     &      0.528  &      0.462  &      0.365     &      0.443     &      0.280  \\
\bottomrule
\end{tabular}\caption[WMT19 Segment-level Results, from English]{WMT19 Segment-level results, metrics (excludes QE as metric results), from English. n denotes number of pairwise judgments.
  \BoldDenotes{ }
  $\baseline$:WMT19 Baseline \cite{ma-etal-2019-results}
  $\submission$:WMT19 Metric Submission \cite{ma-etal-2019-results}
}\label{fig:mt_metric_wmt19_Segment_fromEnglish}
\end{table*}

\begin{table*}[h!]
\centering
\footnotesize
    \addtolength{\tabcolsep}{-2pt} 
\begin{tabular}{lrrrrrrrrrrrrrrrrrrrrrrrrrrrrrrrrrrrrrrrrrrrrrrrrr}
\toprule
                                                                                                           & {\bf de--cs}  & {\bf de--fr} & {\bf fr--de} \\
n                                                                                                          &      35793   &       4862  &       1369 \\
\midrule
\nonen \metric{BEER}${}^\submission$               \cite{stanojevic-simaan-2015-beer}                        &       0.337   &       0.293 &       0.265  \\
\nonen \metric{BERTscore}      \cite{bertscore_arxiv,bert-score}                                           &       0.352   &       0.325 &       0.274  \\
\nonen \metric{CharacTER}${}^\submission$          \cite{wang-etal-2016-character}                           &       0.232  &       0.251 &       0.224 \\
\nonen \metric{chrF}${}^\baseline$               \cite{popovic-2015-chrf}                                    &       0.326  &       0.284 &       0.275 \\
\nonen \metric{chrF+}${}^\baseline$              \cite{popovic-2017-chrf}                                   &       0.326  &       0.284 &       0.278 \\
\nonen \metric{EED}${}^\submission$                \cite{stanchev-etal-2019-eed}                            &       0.345  &       0.301 &       0.267 \\
\nonen \metric{ESIM}${}^\submission$               \cite{chen-etal-2017-enhanced,mathur-etal-2019-putting}  &       0.331  &       0.290 &       0.289 \\
\nonen \metric{hLEPORa\_baseline}${}^\submission$   \cite{han-etal-2012-lepor,han13language}                &       0.207  &       0.239 &        $-$  \\
\nonen \metric{sentBLEU}${}^\baseline$           \cite{papineni-etal-2002-bleu}                             &       0.203  &       0.235 &       0.179 \\
\nonen \metric{YiSi-0}${}^\submission$             \cite{lo-2019-yisi}                                      &       0.331  &       0.296 &       0.277 \\
\nonen \metric{YiSi-1}${}^\submission$             \cite{lo-2019-yisi}                                      &       0.376  &       0.349 &       0.310 \\
\nonen \metric{YiSi-1\_srl}${}^\submission$        \cite{lo-2019-yisi}                                      &        $-$   &       $-$   &       0.299 \\
\midrule
\nonen \BJTwikiplussysref                                                                                  & {\bf 0.458} & {\bf 0.453} & {\bf 0.426}  \\
\nonen \BJTLASERrefLM                                                                                      &      0.431  &      0.401  & {\bf 0.381} \\
\nonen \BJTwikipluslm                                                                                      &      0.294  &      0.235  &      0.138  \\
\nonen \BJTLASERref                                                                                        &      0.397  &      0.352  & {\bf 0.348} \\
\nonen \BJTmBARTsysref                                                                                     &      0.262  &      0.255  &      0.236 \\
\bottomrule
\end{tabular}\caption[WMT19 Segment-level Results, non-English]{WMT19 Segment-level results, metrics (excludes QE as metric), non-English. n denotes number of pairwise judgments. \BoldDenotes{ }$\baseline$:WMT19 Baseline \cite{ma-etal-2019-results} $\submission$:WMT19 Metric Submission \cite{ma-etal-2019-results}
}\label{fig:mt_metric_wmt19_Segment_nonEnglish}
\end{table*}

\begin{table*}[h!]
\centering
\footnotesize
    \addtolength{\tabcolsep}{-2pt} 
\begin{tabular}{lrrrrrrrrrrrrrrrrrrrrrrrrrrrrrrrrrrrrrrrrrrrrrrrrr}
\toprule
                                                                                                           & {\bf de--en}     & {\bf fi--en} & {\bf gu--en}    & {\bf kk--en} & {\bf lt--en} & {\bf ru--en}    & {\bf zh--en} \\
n                                                                                                          &      85365      &      38307  &      31139     &      27094  &      21862  &      46172     &      31070  \\
\midrule
\nonen \metric{ibm1-morpheme}${}^\qeasmetric$      \cite{popovic-etal-2011-evaluation}                     &       -0.074  &      0.009  &         $-$    &         $-$ &      0.069  &         $-$    &         $-$ \\
\nonen \metric{ibm1-pos4gram}${}^\qeasmetric$      \cite{popovic-etal-2011-evaluation}                     &       -0.153  &         $-$ &         $-$    &         $-$ &         $-$ &         $-$    &         $-$ \\
\nonen \metric{LASIM}${}^\qeasmetric$                                                                      &      -0.024   &         $-$ &         $-$    &         $-$ &         $-$ &      0.022     &         $-$ \\
\nonen \metric{LP}${}^\qeasmetric$                                                                         &      -0.096   &         $-$ &         $-$    &         $-$ &         $-$ &      -0.035  &         $-$ \\
\nonen \metric{UNI}${}^\qeasmetric$                \cite{yankovskaya-etal-2019-quality}                    &       0.022     &      0.202  &         $-$    &         $-$ &         $-$ &      0.084     &         $-$ \\
\nonen \metric{UNI+}${}^\qeasmetric$               \cite{yankovskaya-etal-2019-quality}                    &       0.015     &      0.211  &         $-$    &         $-$ &         $-$ &      0.089     &         $-$ \\
\nonen \metric{YiSi-2}${}^\qeasmetric$             \cite{lo-2019-yisi}                                     &       0.068     &      0.126  &      -0.001  &      0.096  &      0.075  &      0.053     &      0.253  \\
\nonen \metric{YiSi-2\_srl}${}^\qeasmetric$        \cite{lo-2019-yisi}                                     &       0.068     &         $-$ &         $-$    &         $-$ &         $-$ &         $-$    &      0.246  \\
\midrule
\nonen \BJTwikiplussysgivensrc                                                                              & {\bf 0.109} & {\bf 0.300} & {\bf 0.102} & {\bf 0.391} & {\bf 0.356} & {\bf 0.178} & {\bf 0.336} \\
\bottomrule
\end{tabular}\caption[WMT19 Segment-level Results, to English]{WMT19 Segment-level results, QE as a metric, to English. n denotes number of pairwise judgments. \BoldDenotes{ }
  $\qeasmetric$:WMT19 QE-as-Metric Submission \cite{fonseca-etal-2019-findings}}\label{fig:mt_metric_wmt19_Segment_toEnglish-qe}
\end{table*}

\begin{table*}[h!]
\centering
\footnotesize
    \addtolength{\tabcolsep}{-2pt} 
\begin{tabular}{lrrrrrrrrrrrrrrrrrrrrrrrrrrrrrrrrrrrrrrrrrrrrrrrrr}
\toprule
                                                                                                           & {\bf en--cs}     & {\bf en--de}    & {\bf en--fi}    & {\bf en--gu} & {\bf en--kk} & {\bf en--lt}    & {\bf en--ru}    & {\bf en--zh} \\
n                                                                                                          &      27178      &      99840     &      31820     &      11355  &      18172  &      17401     &      24334     &      18658  \\
\midrule
\nonen \metric{ibm1-morpheme}${}^\qeasmetric$      \cite{popovic-etal-2011-evaluation}                     &       -0.135  &      -0.003  &      -0.005  &         $-$ &         $-$ &      -0.165  &         $-$    &         $-$ \\
\nonen \metric{ibm1-pos4gram}${}^\qeasmetric$      \cite{popovic-etal-2011-evaluation}                     &          $-$    &      -0.123  &         $-$    &         $-$ &         $-$ &         $-$    &         $-$    &         $-$ \\
\nonen \metric{LASIM}${}^\qeasmetric$                                                                      &         $-$     &      0.147     &         $-$    &         $-$ &         $-$ &         $-$    &      -0.240   &         $-$ \\
\nonen \metric{LP}${}^\qeasmetric$                                                                         &         $-$     &      -0.119  &         $-$    &         $-$ &         $-$ &         $-$    &      -0.158  &         $-$ \\
\nonen \metric{UNI}${}^\qeasmetric$                \cite{yankovskaya-etal-2019-quality}                    &       0.060     &      0.129     &      0.351     &         $-$ &         $-$ &         $-$    &      0.226     &         $-$ \\
\nonen \metric{UNI+}${}^\qeasmetric$               \cite{yankovskaya-etal-2019-quality}                    &          $-$    &         $-$    &         $-$    &         $-$ &         $-$ &         $-$    &      0.222     &         $-$ \\
\nonen \metric{USFD}${}^\qeasmetric$               \cite{ive-etal-2018-deepquest}                          &          $-$    &      -0.029  &         $-$    &         $-$ &         $-$ &         $-$    &      0.136     &         $-$ \\
\nonen \metric{USFD-TL}${}^\qeasmetric$            \cite{ive-etal-2018-deepquest}                          &          $-$    &      -0.037  &         $-$    &         $-$ &         $-$ &         $-$    &      0.191     &         $-$ \\
\nonen \metric{YiSi-2}${}^\qeasmetric$             \cite{lo-2019-yisi}                                     &       0.069     &      0.212     &      0.239     &      0.147  &      0.187  &      0.003     &      -0.155  &      0.044  \\
\nonen \metric{YiSi-2\_srl}${}^\qeasmetric$        \cite{lo-2019-yisi}                                     &          $-$    &      0.236     &         $-$    &         $-$ &         $-$ &         $-$    &         $-$    &      0.034  \\
\midrule
\nonen \BJTwikiplussysgivensrc                                                                             & {\bf 0.470 }     &  {\bf 0.402  } &  {\bf 0.555  } & {\bf 0.215 } &  {\bf 0.507 } & {\bf 0.499 } & {\bf 0.486  } &  {\bf  0.287 } \\
\bottomrule
\end{tabular}\caption[WMT19 Segment-level Results, from English]{WMT19 Segment-level results, QE as a metric, from English. n denotes number of pairwise judgments.
  \BoldDenotes{ }
  $\qeasmetric$:WMT19 QE-as-Metric Submission \cite{fonseca-etal-2019-findings}
}\label{fig:mt_metric_wmt19_Segment_fromEnglish-qe}
\end{table*}

\begin{table*}[h!]
\centering
\footnotesize
    \addtolength{\tabcolsep}{-2pt} 
\begin{tabular}{lrrrrrrrrrrrrrrrrrrrrrrrrrrrrrrrrrrrrrrrrrrrrrrrrr}
\toprule
                                                                                                           & {\bf de--cs}  & {\bf de--fr} & {\bf fr--de} \\
n                                                                                                          &      35793   &       4862  &       1369 \\
\midrule
\nonen \metric{ibm1-morpheme}${}^\qeasmetric$      \cite{popovic-etal-2011-evaluation}                     &       0.048 &       -0.013  &       -0.053 \\
\nonen \metric{ibm1-pos4gram}${}^\qeasmetric$      \cite{popovic-etal-2011-evaluation}                     &        $-$  &       -0.074  &       -0.097 \\
\nonen \metric{YiSi-2}${}^\qeasmetric$             \cite{lo-2019-yisi}                                     &       0.199 &       0.186     &       0.066   \\
\midrule
\nonen \BJTwikiplussysgivensrc                                                                             & {\bf 0.444} &  {\bf 0.371}  &  {\bf 0.316}  \\
\bottomrule
\end{tabular}\caption[WMT19 Segment-level Results, non-English]{WMT19 Segment-level results, QE as a metric, non-English. n denotes number of pairwise judgments. \BoldDenotes{ }
  $\qeasmetric$:WMT19 QE-as-Metric Submission \cite{fonseca-etal-2019-findings}}\label{fig:mt_metric_wmt19_Segment_nonEnglish-qe}
\end{table*}

\clearpage
\section{WMT 2019 System-Level results for Top 4 Systems}\label{appendix:wmt19_top4}

\autoref{fig:mt_metric_wmt19_System_toEnglish_top4}
\autoref{fig:mt_metric_wmt19_System_fromEnglish_top4},
and
\autoref{fig:mt_metric_wmt19_System_nonEnglish_top4}
show system-level results for just the top 4 systems,
for language pairs into, out of, and not including English,
for WMT 2019.
We show statistical significance following the shared task
but note it appears extremely noisy.

\begin{table*}[h!]
\centering
\footnotesize
    \addtolength{\tabcolsep}{-2pt} 
\begin{tabular}{lrrrrrrrrrrrrrrrrrrrrrrrrrrrrrrrrrrrrrrrrrrrrrrrrr}
\toprule
                                                                                                           & {\bf de--en}     & {\bf fi--en}    & {\bf gu--en}    & {\bf kk--en} & {\bf lt--en}    & {\bf ru--en}    & {\bf zh--en} \\
n                                                                                                          &          4      &          4     &          4     &          4  &          4     &          4     &          4  \\
\midrule
\nonen \metric{BEER}${}^\submission$               \cite{stanojevic-simaan-2015-beer}                      &       -0.760   & {\bf 0.065}    & {\bf 0.981}    & {\bf 0.957} & {\bf 0.423}    &      -0.122  &      -0.625  \\
\nonen \metric{BERTr}${}^\submission$              \cite{mathur-etal-2019-putting}                         &       0.251     &      0.430     &      0.966     &      0.864  &      0.518     & {\bf 0.505}    & {\bf 0.402} \\
\nonen \metric{BERTscore}      \cite{bertscore_arxiv,bert-score}                                           &       0.272     & {\bf 0.683}    & {\bf 0.913}    & {\bf 0.897} & {\bf 0.753}    & {\bf 0.456}    &      -0.220  \\
\nonen \metric{BLEU}${}^\baseline$               \cite{papineni-etal-2002-bleu}                            &  {\bf -0.822} & {\bf -0.275} & {\bf 0.966}    & {\bf 0.958} & {\bf 0.625}    & {\bf -0.356} & {\bf -0.694} \\
\nonen \metric{BLEURT}            \cite{bleurt}                                                            & {\bf 0.953}     & {\bf 0.714}    & {\bf 0.881}    &      0.929  & {\bf 0.841}    & {\bf 0.522}    &      0.660  \\
\nonen \metric{CDER}${}^\baseline$               \cite{leusch-etal-2006-cder}                              &       -0.740   & {\bf -0.214} & {\bf 0.940}    &      0.948  &      0.389     & {\bf -0.108} &      -0.611  \\
\nonen \metric{CharacTER}${}^\submission$          \cite{wang-etal-2016-character}                         &  {\bf -0.664} & {\bf -0.079} & {\bf 0.980}    & {\bf 0.924} &      0.386     & {\bf 0.052}    &      -0.092  \\
\nonen \metric{chrF}${}^\baseline$               \cite{popovic-2015-chrf}                                  &       -0.610   & {\bf 0.170}    & {\bf 0.986}    &      0.893  &      0.377     &      -0.043  &      -0.147  \\
\nonen \metric{chrF+}${}^\baseline$              \cite{popovic-2017-chrf}                                  &       -0.612  & {\bf 0.157}    &      0.982     & {\bf 0.886} &      0.341     &      -0.019  &      -0.093  \\
\nonen \metric{EED}${}^\submission$                \cite{stanchev-etal-2019-eed}                           &       -0.503  & {\bf 0.125}    & {\bf 0.978}    & {\bf 0.904} &      0.323     & {\bf 0.033}    &      -0.06  \\
\nonen \metric{ESIM}${}^\submission$               \cite{chen-etal-2017-enhanced,mathur-etal-2019-putting} &       0.895     &      0.740     & {\bf 0.847}    & {\bf 0.965} &      0.896     & {\bf 0.534}    & {\bf 0.819} \\
\nonen \metric{hLEPORa\_baseline}${}^\submission$   \cite{han-etal-2012-lepor,han13language}               &          $-$    &         $-$    &         $-$    &      0.816  &         $-$    &         $-$    & {\bf 0.312} \\
\nonen \metric{hLEPORb\_baseline}${}^\submission$   \cite{han-etal-2012-lepor,han13language}               &          $-$    &         $-$    &         $-$    &      0.816  &      0.257     &         $-$    & {\bf 0.312} \\
\nonen \metric{Meteor++\_2.0(syntax)}${}^\submission$   \cite{guo-hu-2019-meteor}                          &       -0.591  &      0.349     & {\bf 0.978}    & {\bf 0.912} &      0.413     & {\bf 0.024}    &      -0.214  \\
\nonen \metric{Meteor++\_2.0(syntax+copy)}${}^\submission$   \cite{guo-hu-2019-meteor}                     &       -0.587  &      0.399     & {\bf 0.980}    &      0.888  &      0.413     & {\bf 0.051}    &      -0.17  \\
\nonen \metric{NIST}${}^\baseline$               \cite{doddington2002automatic}                            &  {\bf -0.82}  & {\bf 0.111}    & {\bf 0.963}    &      0.913  & {\bf 0.746}    & {\bf -0.458} & {\bf -0.906} \\
\nonen \metric{PER}${}^\baseline$                                                                          & {\bf -0.787}  & {\bf 0.232}    &      0.945     &      0.731  &      0.086     & {\bf -0.081} & {\bf 0.730} \\
\nonen \metric{PReP}${}^\submission$               \cite{yoshimura-etal-2019-filtering}                    &  {\bf -0.981} &      0.754     &      0.976     & {\bf 0.863} &      0.171     & {\bf -0.357} & {\bf -0.927} \\
\nonen \metric{sacreBLEU.BLEU}${}^\baseline$     \cite{post-2018-call}                                     &       -0.823  & {\bf -0.333} & {\bf 0.966}    & {\bf 0.958} &      0.426     & {\bf -0.217} & {\bf -0.694} \\
\nonen \metric{sacreBLEU.chrF}${}^\baseline$     \cite{post-2018-call}                                     &       -0.633  & {\bf 0.113}    &      0.954     & {\bf 0.875} &      0.311     &      -0.094  &      0.347  \\
\nonen \metric{TER}${}^\baseline$                \cite{snover2006study}                                    &       -0.798  & {\bf 0.032}    &      0.942     & {\bf 0.963} & {\bf 0.585}    & {\bf -0.137} &      -0.845  \\
\nonen \metric{WER}${}^\baseline$                                                                          &      -0.816   & {\bf -0.125} & {\bf 0.940}    & {\bf 0.958} & {\bf 0.621}    & {\bf -0.153} & {\bf -0.859} \\
\nonen \metric{WMDO}${}^\submission$               \cite{chow-etal-2019-wmdo}                              &       -0.711  &      0.344     & {\bf 0.943}    & {\bf 0.921} &      0.290     & {\bf 0.114}    &      -0.352  \\
\nonen \metric{YiSi-0}${}^\submission$             \cite{lo-2019-yisi}                                     &       -0.714  & {\bf 0.074}    & {\bf 0.991}    & {\bf 0.946} & {\bf 0.540}    &      -0.079  &      -0.663  \\
\nonen \metric{YiSi-1}${}^\submission$             \cite{lo-2019-yisi}                                     &       0.045     & {\bf 0.610}    & {\bf 0.962}    & {\bf 0.887} & {\bf 0.552}    & {\bf 0.365}    &      -0.067  \\
\nonen \metric{YiSi-1\_srl}${}^\submission$        \cite{lo-2019-yisi}                                     &       0.081     & {\bf 0.580}    & {\bf 0.959}    & {\bf 0.874} & {\bf 0.560}    &      0.342     &      -0.069  \\
\midrule
\nonen \metric{ibm1-morpheme}${}^\qeasmetric$      \cite{popovic-etal-2011-evaluation}                     &  {\bf -0.643} & {\bf 0.065}    &         $-$    &         $-$ & {\bf -0.952} &         $-$    &         $-$ \\
\nonen \metric{ibm1-pos4gram}${}^\qeasmetric$      \cite{popovic-etal-2011-evaluation}                     &  {\bf -0.831} &         $-$    &         $-$    &         $-$ &         $-$    &         $-$    &         $-$ \\
\nonen \metric{LASIM}${}^\qeasmetric$                                                                      & {\bf -0.855}  &         $-$    &         $-$    &         $-$ &         $-$    & {\bf -0.353} &         $-$ \\
\nonen \metric{LP.1}${}^\qeasmetric$                                                                       & {\bf 0.777}     &         $-$    &         $-$    &         $-$ &         $-$    &      0.442     &         $-$ \\
\nonen \metric{UNI}${}^\qeasmetric$                \cite{yankovskaya-etal-2019-quality}                    &  {\bf 0.703}    & {\bf 0.830}    &         $-$    &         $-$ &         $-$    &      0.738     &         $-$ \\
\nonen \metric{UNI+}${}^\qeasmetric$               \cite{yankovskaya-etal-2019-quality}                    &  {\bf 0.796}    & {\bf 0.791}    &         $-$    &         $-$ &         $-$    & {\bf 0.777}    &         $-$ \\
\nonen \metric{YiSi-2}${}^\qeasmetric$             \cite{lo-2019-yisi}                                     &  {\bf -0.809} & {\bf 0.780}    & {\bf -0.125} & {\bf 0.834} & {\bf -0.362} & {\bf -0.325} & {\bf -0.889} \\
\nonen \metric{YiSi-2\_srl}${}^\qeasmetric$        \cite{lo-2019-yisi}                                     &  {\bf -0.749} &         $-$    &         $-$    &         $-$ &         $-$    &         $-$    &      -0.83  \\
\midrule
\nonen \BJTwikiplussysref                                                                                  &      0.401      & {\bf 0.719}    & {\bf 0.896}    &      0.796  & {\bf 0.877}    & {\bf 0.431}    &      0.523  \\
\nonen \BJTParabanksysref                                                                                  & {\bf 0.957}     & {\bf 0.788}    & {\bf 0.871}    & {\bf 0.759} & {\bf 0.939}    & {\bf 0.625}    & {\bf 0.899} \\
\nonen \BJTLASERrefLM                                                                                      & {\bf 0.957}     &      0.768     &      0.867     & {\bf 0.870} &      0.615     & {\bf 0.596}    & {\bf 0.733} \\
\nonen \BJTwikiplussysgivensrc                                                                             &      0.502      & {\bf 0.802}    &      0.608     &      0.558  & {\bf -0.301} & {\bf 0.437}    & {\bf 0.958} \\
\nonen \BJTwikipluslm                                                                                      & {\bf 0.973}     & {\bf 0.754}    &      0.619     & {\bf 0.498} & {\bf -0.006} & {\bf 0.779}    & {\bf 0.973} \\
\nonen \BJTLASERref                                                                                        &      -0.458   & {\bf 0.718}    & {\bf 0.984}    &      0.926  & {\bf 0.662}    & {\bf 0.262}    &      -0.528  \\
\nonen \BJTmBARTsysref                                                                                     &      -0.739   & {\bf 0.559}    &      0.913     & {\bf 0.902} &      0.491     &      -0.103  &      -0.295  \\
\bottomrule
\end{tabular}\caption[WMT19 System-level Results, to English]{WMT19 System-level results, to English for the top 4 systems (as judged by humans) for each language pair. n denotes number of MT systems. \BoldDenotes{ }$\baseline$:WMT19 Baseline \cite{ma-etal-2019-results} $\submission$:WMT19 Metric Submission \cite{ma-etal-2019-results} $\qeasmetric$:WMT19 QE-as-Metric Submission \cite{fonseca-etal-2019-findings}}\label{fig:mt_metric_wmt19_System_toEnglish_top4}
\end{table*}

\begin{table*}[h!]
\centering
\footnotesize
    \addtolength{\tabcolsep}{-3pt} 
\begin{tabular}{lrrrrrrrrrrrrrrrrrrrrrrrrrrrrrrrrrrrrrrrrrrrrrrrrr}
\toprule
                                                                                                           & {\bf en--cs}     & {\bf en--de}    & {\bf en--fi}    & {\bf en--gu}    & {\bf en--kk}    & {\bf en--lt}    & {\bf en--ru}    & {\bf en--zh} \\
n                                                                                                          &          4      &          4     &          4     &          4     &          4     &          4     &          4     &          4  \\
\midrule
\nonen \metric{BEER}${}^\submission$               \cite{stanojevic-simaan-2015-beer}                      &       0.872     & {\bf -0.801} & {\bf 0.960}    &      0.899     &      0.226     & {\bf 0.888}    & {\bf 0.961}    & {\bf 0.992} \\
\nonen \metric{BERTscore}      \cite{bertscore_arxiv,bert-score}                                           &  {\bf 0.868}    &      -0.722  & {\bf 0.859}    & {\bf 0.922}    &      0.288     & {\bf 0.955}    & {\bf 0.953}    &      0.982  \\
\nonen \metric{BLEU}${}^\baseline$               \cite{papineni-etal-2002-bleu}                            &       0.930     & {\bf -0.37}  &      0.898     & {\bf 0.860}    &      0.181     & {\bf 0.925}    &      0.753     &      0.987  \\
\nonen \metric{CDER}${}^\baseline$               \cite{leusch-etal-2006-cder}                              &  {\bf 0.946}    & {\bf -0.975} &      0.837     & {\bf 0.900}    & {\bf -0.011} & {\bf 0.880}    & {\bf 0.917}    & {\bf 0.986} \\
\nonen \metric{CharacTER}${}^\submission$          \cite{wang-etal-2016-character}                         &       0.828     & {\bf -0.777} & {\bf 0.887}    & {\bf 0.902}    &      0.295     & {\bf 0.675}    & {\bf 0.974}    & {\bf 0.997} \\
\nonen \metric{chrF}${}^\baseline$               \cite{popovic-2015-chrf}                                  &       0.799     &      -0.590   & {\bf 0.936}    & {\bf 0.926}    &      0.277     & {\bf 0.901}    & {\bf 0.954}    &      0.987  \\
\nonen \metric{chrF+}${}^\baseline$              \cite{popovic-2017-chrf}                                  &       0.816     &      -0.605  & {\bf 0.921}    & {\bf 0.923}    &      0.283     & {\bf 0.858}    &      0.940     & {\bf 0.996} \\
\nonen \metric{EED}${}^\submission$                \cite{stanchev-etal-2019-eed}                           &       0.825     &      -0.552  & {\bf 0.939}    & {\bf 0.913}    &      0.267     & {\bf 0.921}    &      0.961     & {\bf 0.997} \\
\nonen \metric{ESIM}${}^\submission$               \cite{chen-etal-2017-enhanced,mathur-etal-2019-putting} &          $-$    & {\bf -0.796} & {\bf 0.957}    &         $-$    &      0.418     & {\bf 0.997}    & {\bf 0.986}    &      0.987  \\
\nonen \metric{hLEPORa\_baseline}${}^\submission$   \cite{han-etal-2012-lepor,han13language}               &          $-$    &         $-$    &         $-$    &      0.915     & {\bf 0.062}    &         $-$    &         $-$    &         $-$ \\
\nonen \metric{hLEPORb\_baseline}${}^\submission$   \cite{han-etal-2012-lepor,han13language}               &          $-$    &         $-$    &         $-$    & {\bf 0.915}    & {\bf 0.062}    &      0.821     &         $-$    &         $-$ \\
\nonen \metric{NIST}${}^\baseline$               \cite{doddington2002automatic}                            &  {\bf 0.946}    & {\bf -0.233} & {\bf 0.971}    & {\bf 0.893}    &      0.082     & {\bf 0.988}    &      0.724     &      0.979  \\
\nonen \metric{PER}${}^\baseline$                                                                          &      0.916      & {\bf -0.995} & {\bf 0.850}    & {\bf 0.887}    & {\bf -0.260}  &      0.390     & {\bf 0.911}    & {\bf 0.980} \\
\nonen \metric{sacreBLEU.BLEU}${}^\baseline$     \cite{post-2018-call}                                     &  {\bf 0.970}    & {\bf -0.976} &      0.845     &      0.859     &      0.181     &      0.638     & {\bf 0.878}    &      0.962  \\
\nonen \metric{sacreBLEU.chrF}${}^\baseline$     \cite{post-2018-call}                                     &       0.907     & {\bf -0.816} & {\bf 0.921}    & {\bf 0.902}    &      0.239     & {\bf 0.980}    & {\bf 0.970}    & {\bf 0.963} \\
\nonen \metric{TER}${}^\baseline$                \cite{snover2006study}                                    &  {\bf 0.969}    & {\bf -0.989} & {\bf 0.889}    &      0.874     & {\bf -0.060}  & {\bf 0.988}    & {\bf 0.895}    &      0.984  \\
\nonen \metric{WER}${}^\baseline$                                                                          & {\bf 0.973}     & {\bf -0.993} &      0.876     &      0.868     & {\bf -0.058} & {\bf 0.973}    & {\bf 0.894}    &      0.987  \\
\nonen \metric{YiSi-0}${}^\submission$             \cite{lo-2019-yisi}                                     &       0.879     &      -0.796  & {\bf 0.975}    & {\bf 0.920}    &      0.196     &      0.787     & {\bf 0.940}    &      0.982  \\
\nonen \metric{YiSi-1}${}^\submission$             \cite{lo-2019-yisi}                                     &       0.847     &      -0.220  & {\bf 0.976}    & {\bf 0.917}    &      0.342     & {\bf 0.838}    & {\bf 0.963}    & {\bf 0.990} \\
\nonen \metric{YiSi-1\_srl}${}^\submission$        \cite{lo-2019-yisi}                                     &          $-$    &      -0.378  &         $-$    &         $-$    &         $-$    &         $-$    &         $-$    & {\bf 0.994} \\
\midrule
\nonen \metric{ibm1-morpheme}${}^\qeasmetric$      \cite{popovic-etal-2011-evaluation}                     &  {\bf -0.771} & {\bf -0.425} & {\bf 0.430}    &         $-$    &         $-$    & {\bf 0.969}    &         $-$    &         $-$ \\
\nonen \metric{ibm1-pos4gram}${}^\qeasmetric$      \cite{popovic-etal-2011-evaluation}                     &          $-$    &      -0.502  &         $-$    &         $-$    &         $-$    &         $-$    &         $-$    &         $-$ \\
\nonen \metric{LASIM}${}^\qeasmetric$                                                                      &         $-$     &      -0.914  &         $-$    &         $-$    &         $-$    &         $-$    &      0.223     &         $-$ \\
\nonen \metric{LP.1}${}^\qeasmetric$                                                                       &         $-$     &      0.949     &         $-$    &         $-$    &         $-$    &         $-$    &      -0.407  &         $-$ \\
\nonen \metric{UNI}${}^\qeasmetric$                \cite{yankovskaya-etal-2019-quality}                    &       0.587     & {\bf -0.96}  &      0.637     &         $-$    &         $-$    &         $-$    & {\bf 0.655}    &         $-$ \\
\nonen \metric{UNI+}${}^\qeasmetric$               \cite{yankovskaya-etal-2019-quality}                    &          $-$    &         $-$    &         $-$    &         $-$    &         $-$    &         $-$    &      0.644     &         $-$ \\
\nonen \metric{USFD}${}^\qeasmetric$               \cite{ive-etal-2018-deepquest}                          &          $-$    & {\bf -0.729} &         $-$    &         $-$    &         $-$    &         $-$    & {\bf 0.985}    &         $-$ \\
\nonen \metric{USFD-TL}${}^\qeasmetric$            \cite{ive-etal-2018-deepquest}                          &          $-$    &      -0.390   &         $-$    &         $-$    &         $-$    &         $-$    &      0.698     &         $-$ \\
\nonen \metric{YiSi-2}${}^\qeasmetric$             \cite{lo-2019-yisi}                                     &  {\bf 0.793}    &      -0.933  & {\bf -0.991} &      -0.389  & {\bf 0.851}    &      -0.504  &      0.075     &      0.983  \\
\nonen \metric{YiSi-2\_srl}${}^\qeasmetric$        \cite{lo-2019-yisi}                                     &          $-$    &      -0.915  &         $-$    &         $-$    &         $-$    &         $-$    &         $-$    & {\bf 0.991} \\
\midrule
\nonen \BJTwikiplussysref                                                                                  & {\bf 0.952}     & {\bf 0.278}    &      0.886     & {\bf 0.863}    &      0.693     &      0.862     & {\bf 0.975}    & {\bf 0.966} \\
\nonen \BJTLASERrefLM                                                                                      & {\bf 0.961}     & {\bf 0.377}    &      0.903     & {\bf 0.509}    &      0.605     & {\bf 0.743}    &      0.962     & {\bf 0.985} \\
\nonen \BJTwikiplussysgivensrc                                                                             & {\bf 0.973}     &      -0.408  &      0.765     & {\bf -0.703} & {\bf 0.833}    &      -0.003  &      0.708     & {\bf 0.863} \\
\nonen \BJTwikipluslm                                                                                      & {\bf 0.833}     & {\bf 0.425}    &      0.763     & {\bf -0.712} & {\bf 0.953}    & {\bf 0.633}    &      0.916     & {\bf 0.846} \\
\nonen \BJTLASERref                                                                                        &      0.851      & {\bf 0.246}    & {\bf 0.983}    & {\bf 0.568}    & {\bf 0.328}    & {\bf 0.263}    & {\bf 0.995}    &      0.988  \\
\nonen \BJTmBARTsysref                                                                                     &      0.936      &      -0.834  & {\bf 0.966}    &      0.912     & {\bf 0.224}    & {\bf 0.946}    & {\bf 0.968}    & {\bf 0.986} \\
\bottomrule
\end{tabular}\caption[WMT19 System-level Results, from English]{WMT19 System-level results, from English for the top 4 systems (as judged by humans) for each language pair. n denotes number of MT systems. \BoldDenotes{ }$\baseline$:WMT19 Baseline \cite{ma-etal-2019-results} $\submission$:WMT19 Metric Submission \cite{ma-etal-2019-results} $\qeasmetric$:WMT19 QE-as-Metric Submission \cite{fonseca-etal-2019-findings}}\label{fig:mt_metric_wmt19_System_fromEnglish_top4}
\end{table*}

\begin{table*}[h!]
\centering
\footnotesize
    \addtolength{\tabcolsep}{-2pt} 
\begin{tabular}{lrrrrrrrrrrrrrrrrrrrrrrrrrrrrrrrrrrrrrrrrrrrrrrrrr}
\toprule
                                                                                                           & {\bf de--cs}       & {\bf de--fr}      & {\bf fr--de}        \\
n                                                                                                          &          4        &          4       &          4          \\
\midrule                                                                                                                                                                
\nonen \metric{BEER}${}^\submission$               \cite{stanojevic-simaan-2015-beer}                      &  {\bf 0.961}      &  {\bf 0.590}     &       0.978          \\
\nonen \metric{BERTscore}      \cite{bertscore_arxiv,bert-score}                                           &  {\bf 0.976}      &  {\bf 0.707}     &  {\bf 0.973}         \\
\nonen \metric{BLEU}${}^\baseline$               \cite{papineni-etal-2002-bleu}                            &       0.812       &       0.495      &  {\bf 0.983}         \\
\nonen \metric{CDER}${}^\baseline$               \cite{leusch-etal-2006-cder}                              &  {\bf 0.860}      &       0.544      &       0.959          \\
\nonen \metric{CharacTER}${}^\submission$          \cite{wang-etal-2016-character}                         &  {\bf 0.871}      &  {\bf 0.626}     &       0.963          \\
\nonen \metric{chrF}${}^\baseline$               \cite{popovic-2015-chrf}                                  &  {\bf 0.920}      &       0.531      &  {\bf 0.952}         \\
\nonen \metric{chrF+}${}^\baseline$              \cite{popovic-2017-chrf}                                  &       0.909       &       0.522      &       0.946          \\
\nonen \metric{EED}${}^\submission$                \cite{stanchev-etal-2019-eed}                           &  {\bf 0.873}      &       0.582      &       0.945          \\
\nonen \metric{ESIM}${}^\submission$               \cite{chen-etal-2017-enhanced,mathur-etal-2019-putting} &  {\bf 0.977}      &  {\bf 0.702}     &  {\bf 0.991}         \\
\nonen \metric{hLEPORa\_baseline}${}^\submission$   \cite{han-etal-2012-lepor,han13language}               &       0.771       &       0.314                             \\
\nonen \metric{hLEPORb\_baseline}${}^\submission$   \cite{han-etal-2012-lepor,han13language}               &  {\bf 0.754}      &       0.314                             \\
\nonen \metric{NIST}${}^\baseline$               \cite{doddington2002automatic}                            &       0.754       &       0.561      &  {\bf 0.990}         \\
\nonen \metric{PER}${}^\baseline$                                                                          &      0.913        &      0.401       & {\bf 0.990}         \\
\nonen \metric{sacreBLEU.BLEU}${}^\baseline$     \cite{post-2018-call}                                     &  {\bf 0.888}      &       0.495      &       0.958          \\
\nonen \metric{sacreBLEU.chrF}${}^\baseline$     \cite{post-2018-call}                                     &  {\bf 0.964}      &       0.575      &       0.920          \\
\nonen \metric{TER}${}^\baseline$                \cite{snover2006study}                                    &  {\bf 0.999}      &       0.541      &  {\bf 0.989}         \\
\nonen \metric{WER}${}^\baseline$                                                                          & {\bf 0.997}       & {\bf 0.566}      & {\bf 0.991}         \\
\nonen \metric{YiSi-0}${}^\submission$             \cite{lo-2019-yisi}                                     &       0.838       &  {\bf 0.655}     &       0.961          \\
\nonen \metric{YiSi-1}${}^\submission$             \cite{lo-2019-yisi}                                     &  {\bf 0.967}      &       0.677      &       0.967          \\
\nonen \metric{YiSi-1\_srl}${}^\submission$        \cite{lo-2019-yisi}                                     &    $-$            &      $-$         &  {\bf 0.974}         \\
\midrule                                                                                                                           
\nonen \metric{ibm1-morpheme}${}^\qeasmetric$      \cite{popovic-etal-2011-evaluation}                     &       0.645       &  {\bf -0.885}  &  {\bf -0.339}      \\
\nonen \metric{ibm1-pos4gram}${}^\qeasmetric$      \cite{popovic-etal-2011-evaluation}                     &     $-$           &  {\bf -0.106}  &       -0.33        \\
\nonen \metric{YiSi-2}${}^\qeasmetric$             \cite{lo-2019-yisi}                                     &       0.368       &       0.209      &  {\bf -0.687}      \\
\midrule                                                                                                                                                                
\nonen \BJTwikiplussysref                                                                                  & {\bf 0.968}       & {\bf 0.648}      & {\bf 0.998}         \\
\nonen \BJTLASERrefLM                                                                                      & {\bf 0.947}       & {\bf 0.774}      &      0.975          \\
\nonen \BJTwikiplussysgivensrc                                                                             & {\bf 0.903}       & {\bf 0.600}      & {\bf 0.181}         \\
\nonen \BJTwikipluslm                                                                                      &      0.336        & {\bf 0.770}      &   -0.903          \\
\nonen \BJTLASERref                                                                                        & {\bf 0.552}       & {\bf 0.713}      &      0.953          \\
\nonen \BJTmBARTsysref                                                                                     &      0.806        & {\bf 0.615}      & {\bf 0.972}         \\
\bottomrule
\end{tabular}\caption[WMT19 System-level Results, non-English]{WMT19 System-level results, non-English for the top 4 systems (as judged by humans) for each language pair. n denotes number of MT systems. \BoldDenotes{ }$\baseline$:WMT19 Baseline \cite{ma-etal-2019-results} $\submission$:WMT19 Metric Submission \cite{ma-etal-2019-results} $\qeasmetric$:WMT19 QE-as-Metric Submission \cite{fonseca-etal-2019-findings}}\label{fig:mt_metric_wmt19_System_nonEnglish_top4}
\end{table*}

\clearpage
\section{WMT 2019 Metric and QE as Metric System-Level Results}\label{appendix:wmt19_sys}

\autoref{fig:mt_metric_wmt19_System_toEnglish},
\autoref{fig:mt_metric_wmt19_System_fromEnglish},
and
\autoref{fig:mt_metric_wmt19_System_nonEnglish},
show system-level results, for metrics (excludes QE as metric)
for language pairs into, out of, and not including English,
for the WMT 2019 MT metrics shared task, along with
all baselines and submitted systems.

\autoref{fig:mt_metric_wmt19_System_toEnglish-qe},
\autoref{fig:mt_metric_wmt19_System_fromEnglish-qe},
and
\autoref{fig:mt_metric_wmt19_System_nonEnglish-qe},
show system-level results, for QE as metric,
for language pairs into, out of, and not including English,
for the WMT 2019 MT metrics shared task, along with
all baselines and submitted systems. 

\begin{table*}[h!]
\centering
\footnotesize
    \addtolength{\tabcolsep}{-2pt} 
\begin{tabular}{lrrrrrrrrrrrrrrrrrrrrrrrrrrrrrrrrrrrrrrrrrrrrrrrrr}
\toprule
                                                                                                           & {\bf de--en}     & {\bf fi--en} & {\bf gu--en}    & {\bf kk--en}    & {\bf lt--en} & {\bf ru--en}    & {\bf zh--en} \\
n                                                                                                          &         16      &         12  &         11     &         11     &         11  &         14     &         15  \\
\midrule
\nonen \metric{BEER}${}^\submission$               \cite{stanojevic-simaan-2015-beer}                      &       0.906     & {\bf 0.993} &      0.952     &      0.986     &      0.947  &      0.915     &      0.942  \\
\nonen \metric{BERTr}${}^\submission$              \cite{mathur-etal-2019-putting}                         &  {\bf 0.926}    &      0.984  &      0.938     &      0.990     &      0.948  & {\bf 0.971}    &      0.974  \\
\nonen \metric{BERTscore}      \cite{bertscore_arxiv,bert-score}                                           &  {\bf 0.949}    & {\bf 0.987} & {\bf 0.981}    &      0.980     &      0.962  &      0.921     &      0.983  \\
\nonen \metric{BLEU}${}^\baseline$               \cite{papineni-etal-2002-bleu}                            &       0.849     &      0.982  &      0.834     &      0.946     &      0.961  &      0.879     &      0.899  \\
\nonen \metric{BLEURT}        \cite{bleurt}                                                                & {\bf 0.940}     &      0.978  &      0.878     &      0.993     & {\bf 0.991} & {\bf 0.977}    &      0.984  \\
\nonen \metric{CDER}${}^\baseline$               \cite{leusch-etal-2006-cder}                              &       0.890     & {\bf 0.988} &      0.876     &      0.967     &      0.975  &      0.892     &      0.917  \\
\nonen \metric{CharacTER}${}^\submission$          \cite{wang-etal-2016-character}                         &       0.898     & {\bf 0.990} &      0.922     &      0.953     &      0.955  &      0.923     &      0.943  \\
\nonen \metric{chrF}${}^\baseline$               \cite{popovic-2015-chrf}                                  &  {\bf 0.917}    & {\bf 0.992} &      0.955     &      0.978     &      0.940  &      0.945     &      0.956  \\
\nonen \metric{chrF+}${}^\baseline$              \cite{popovic-2017-chrf}                                  &  {\bf 0.916}    & {\bf 0.992} &      0.947     &      0.976     &      0.940  &      0.945     &      0.956  \\
\nonen \metric{EED}${}^\submission$                \cite{stanchev-etal-2019-eed}                           &       0.903     & {\bf 0.994} &      0.976     &      0.980     &      0.929  &      0.950     &      0.949  \\
\nonen \metric{ESIM}${}^\submission$               \cite{chen-etal-2017-enhanced,mathur-etal-2019-putting} &  {\bf 0.941}    &      0.971  &      0.885     &      0.986     & {\bf 0.989} & {\bf 0.968}    & {\bf 0.988} \\
\nonen \metric{hLEPORa\_baseline}${}^\submission$   \cite{han-etal-2012-lepor,han13language}               &          $-$    &         $-$ &         $-$    &      0.975     &         $-$ &         $-$    &      0.947  \\
\nonen \metric{hLEPORb\_baseline}${}^\submission$   \cite{han-etal-2012-lepor,han13language}               &          $-$    &         $-$ &         $-$    &      0.975     &      0.906  &         $-$    &      0.947  \\
\nonen \metric{Meteor++\_2.0(syntax)}${}^\submission$   \cite{guo-hu-2019-meteor}                          &       0.887     & {\bf 0.995} &      0.909     &      0.974     &      0.928  & {\bf 0.950}    &      0.948  \\
\nonen \metric{Meteor++\_2.0(syntax+copy)}${}^\submission$   \cite{guo-hu-2019-meteor}                     &       0.896     & {\bf 0.995} &      0.900     &      0.971     &      0.927  & {\bf 0.952}    &      0.952  \\
\nonen \metric{NIST}${}^\baseline$               \cite{doddington2002automatic}                            &       0.813     &      0.986  &      0.930     &      0.942     &      0.944  &      0.925     &      0.921  \\
\nonen \metric{PER}${}^\baseline$                                                                          &      0.883      & {\bf 0.991} &      0.910     &      0.737     &      0.947  &      0.922     &      0.952  \\
\nonen \metric{PReP}${}^\submission$               \cite{yoshimura-etal-2019-filtering}                    &       0.575     &      0.614  &      0.773     &      0.776     &      0.494  &      0.782     &      0.592  \\
\nonen \metric{sacreBLEU.BLEU}${}^\baseline$     \cite{post-2018-call}                                     &       0.813     &      0.985  &      0.834     &      0.946     &      0.955  &      0.873     &      0.903  \\
\nonen \metric{sacreBLEU.chrF}${}^\baseline$     \cite{post-2018-call}                                     &       0.910     & {\bf 0.990} &      0.952     &      0.969     &      0.935  &      0.919     &      0.955  \\
\nonen \metric{TER}${}^\baseline$                \cite{snover2006study}                                    &       0.874     & {\bf 0.984} &      0.890     &      0.799     &      0.960  &      0.917     &      0.840  \\
\nonen \metric{WER}${}^\baseline$                                                                          &      0.863      &      0.983  &      0.861     &      0.793     &      0.961  &      0.911     &      0.820  \\
\nonen \metric{WMDO}${}^\submission$               \cite{chow-etal-2019-wmdo}                              &       0.872     & {\bf 0.987} &      0.983     & {\bf 0.998}    &      0.900  &      0.942     &      0.943  \\
\nonen \metric{YiSi-0}${}^\submission$             \cite{lo-2019-yisi}                                     &       0.902     & {\bf 0.993} & {\bf 0.993}    &      0.991     &      0.927  & {\bf 0.958}    &      0.937  \\
\nonen \metric{YiSi-1}${}^\submission$             \cite{lo-2019-yisi}                                     &  {\bf 0.949}    & {\bf 0.989} &      0.924     &      0.994     &      0.981  & {\bf 0.979}    &      0.979  \\
\nonen \metric{YiSi-1\_srl}${}^\submission$        \cite{lo-2019-yisi}                                     &  {\bf 0.950}    & {\bf 0.989} &      0.918     &      0.994     &      0.983  & {\bf 0.978}    &      0.977  \\
\midrule
\nonen \BJTwikiplussysref                                                                                  & {\bf 0.954}     &      0.983  &      0.764     & {\bf 0.998}    & {\bf 0.995} &      0.914     & {\bf 0.992} \\
\nonen \BJTParabanksysref                                                                                  & {\bf 0.949}     &      0.979  &      0.925     &      0.993     &      0.981  &      0.948     & {\bf 0.994} \\
\nonen \BJTLASERrefLM                                                                                      & {\bf 0.938}     &      0.974  &      0.974     & {\bf 0.997}    & {\bf 0.996} &      0.940     &      0.988  \\
\nonen \BJTmBARTsysref                                                                                      &      0.906     & {\bf 0.991} &         0.949  &         0.974  &      0.917  &      0.880     &      0.956  \\
\bottomrule
\end{tabular}\caption[WMT19 System-level Results, to English]{WMT19 System-level results, to English. 
n denotes number of MT systems. \BoldDenotes{ }
$\baseline$:WMT19 Baseline \cite{ma-etal-2019-results} 
$\submission$:WMT19 Metric Submission \cite{ma-etal-2019-results} 
}\label{fig:mt_metric_wmt19_System_toEnglish}
\end{table*}

\begin{table*}[h!]
\centering
\footnotesize
    \addtolength{\tabcolsep}{-2pt} 
\begin{tabular}{lrrrrrrrrrrrrrrrrrrrrrrrrrrrrrrrrrrrrrrrrrrrrrrrrr}
\toprule
                                                                                                           & {\bf en--cs}     & {\bf en--de}    & {\bf en--fi} & {\bf en--gu} & {\bf en--kk} & {\bf en--lt}   & {\bf en--ru}    & {\bf en--zh} \\
n                                                                                                          &         11      &         22     &         12  &         11  &         11  &         12    &         12     &         12  \\
\midrule
\nonen \metric{BEER}${}^\submission$               \cite{stanojevic-simaan-2015-beer}                      &  {\bf 0.990}    &      0.983     & {\bf 0.989} &      0.829  &      0.971  & {\bf 0.982}   &      0.977     &      0.803  \\
\nonen \metric{BERTscore}      \cite{bertscore_arxiv,bert-score}                                           &       0.981     & {\bf 0.990}    &      0.970  & {\bf 0.922} &      0.981  &      0.978    & {\bf 0.989}    &      0.925  \\
\nonen \metric{BLEU}${}^\baseline$               \cite{papineni-etal-2002-bleu}                            &       0.897     &      0.921     & {\bf 0.969} &      0.737  &      0.852  & {\bf 0.989}   &      0.986     &      0.901  \\
\nonen \metric{CDER}${}^\baseline$               \cite{leusch-etal-2006-cder}                              &       0.985     &      0.973     & {\bf 0.978} &      0.840  &      0.927  & {\bf 0.985}   & {\bf 0.993}    &      0.905  \\
\nonen \metric{CharacTER}${}^\submission$          \cite{wang-etal-2016-character}                         &  {\bf 0.994}    & {\bf 0.986}    &      0.968  & {\bf 0.910} &      0.936  &      0.954    & {\bf 0.985}    &      0.862  \\
\nonen \metric{chrF}${}^\baseline$               \cite{popovic-2015-chrf}                                  &       0.990     &      0.979     & {\bf 0.986} & {\bf 0.841} & {\bf 0.972} & {\bf 0.981}   &      0.943     &      0.880  \\
\nonen \metric{chrF+}${}^\baseline$              \cite{popovic-2017-chrf}                                  &  {\bf 0.991}    &      0.981     & {\bf 0.986} &      0.848  & {\bf 0.974} & {\bf 0.982}   &      0.950     &      0.879  \\
\nonen \metric{EED}${}^\submission$                \cite{stanchev-etal-2019-eed}                           &  {\bf 0.993}    & {\bf 0.985}    & {\bf 0.987} &      0.897  & {\bf 0.979} &      0.975    &      0.967     &      0.856  \\
\nonen \metric{ESIM}${}^\submission$               \cite{chen-etal-2017-enhanced,mathur-etal-2019-putting} &          $-$    & {\bf 0.991}    &      0.957  &         $-$ & {\bf 0.980} & {\bf 0.989}   & {\bf 0.989}    &      0.931  \\
\nonen \metric{hLEPORa\_baseline}${}^\submission$   \cite{han-etal-2012-lepor,han13language}               &          $-$    &         $-$    &         $-$ &      0.841  &      0.968  &         $-$   &         $-$    &         $-$ \\
\nonen \metric{hLEPORb\_baseline}${}^\submission$   \cite{han-etal-2012-lepor,han13language}               &          $-$    &         $-$    &         $-$ &      0.841  &      0.968  &      0.980    &         $-$    &         $-$ \\
\nonen \metric{NIST}${}^\baseline$               \cite{doddington2002automatic}                            &       0.896     &      0.321     &      0.971  &      0.786  &      0.930  & {\bf 0.993}   & {\bf 0.988}    &      0.884  \\
\nonen \metric{PER}${}^\baseline$                                                                          &      0.976      &      0.970     & {\bf 0.982} &      0.839  &      0.921  &      0.985    &      0.981     &      0.895  \\
\nonen \metric{sacreBLEU.BLEU}${}^\baseline$     \cite{post-2018-call}                                     &  {\bf 0.994}    &      0.969     &      0.966  &      0.736  &      0.852  & {\bf 0.986}   &      0.977     &      0.801  \\
\nonen \metric{sacreBLEU.chrF}${}^\baseline$     \cite{post-2018-call}                                     &       0.983     &      0.976     &      0.980  &      0.841  & {\bf 0.967} &      0.966    & {\bf 0.985}    &      0.796  \\
\nonen \metric{TER}${}^\baseline$                \cite{snover2006study}                                    &       0.980     &      0.969     & {\bf 0.981} & {\bf 0.865} &      0.940  & {\bf 0.994}   & {\bf 0.995}    &      0.856  \\
\nonen \metric{WER}${}^\baseline$                                                                          &      0.982      &      0.966     & {\bf 0.980} & {\bf 0.861} &      0.939  & {\bf 0.991}   & {\bf 0.994}    &      0.875  \\
\nonen \metric{YiSi-0}${}^\submission$             \cite{lo-2019-yisi}                                     &  {\bf 0.992}    &      0.985     & {\bf 0.987} &      0.863  &      0.974  &      0.974    &      0.953     &      0.861  \\
\nonen \metric{YiSi-1}${}^\submission$             \cite{lo-2019-yisi}                                     &       0.962     & {\bf 0.991}    &      0.971  & {\bf 0.909} & {\bf 0.985} &      0.963    & {\bf 0.992}    & {\bf 0.951} \\
\nonen \metric{YiSi-1\_srl}${}^\submission$        \cite{lo-2019-yisi}                                     &          $-$    & {\bf 0.991}    &         $-$ &         $-$ &         $-$ &         $-$   &         $-$    & {\bf 0.948} \\
\midrule
\nonen \BJTwikiplussysref                                                                                  &      0.958      & {\bf 0.988}    &      0.949  &      0.624  & {\bf 0.978} &      0.937    &      0.918     &      0.898  \\
\nonen \BJTLASERrefLM                                                                                      &      0.962      & {\bf 0.989}    &      0.957  &      0.775  &      0.969  &      0.958    & {\bf 0.987}    & {\bf 0.950} \\
\nonen \BJTmBARTsysref                                                                                    & {\bf 0.987}     &      0.988     & {\bf 0.982} & {\bf 0.917} & {\bf 0.981} &      0.965    &      0.978     &      0.866  \\
\bottomrule
\end{tabular}\caption[WMT19 System-level Results, from English]{WMT19 System-level results, from English. 
n denotes number of MT systems. \BoldDenotes{ }
$\baseline$:WMT19 Baseline \cite{ma-etal-2019-results} $\submission$:WMT19 Metric Submission \cite{ma-etal-2019-results} 
}\label{fig:mt_metric_wmt19_System_fromEnglish}
\end{table*}

\begin{table*}[h!]
\centering
\footnotesize
    \addtolength{\tabcolsep}{-2pt} 
\begin{tabular}{lrrrrrrrrrrrrrrrrrrrrrrrrrrrrrrrrrrrrrrrrrrrrrrrrr}
\toprule
                                                                                                           & {\bf de--cs}  & {\bf de--fr}  & {\bf fr--de}  \\
n                                                                                                          &         11   &         11   &         10   \\
\midrule                                                                                                                                 
\nonen \metric{BEER}${}^\submission$               \cite{stanojevic-simaan-2015-beer}                      &  {\bf 0.978} &  {\bf 0.941}  &       0.848  \\
\nonen \metric{BERTscore}      \cite{bertscore_arxiv,bert-score}                                           &       0.969  &  {\bf 0.971}  &  {\bf 0.899} \\
\nonen \metric{BLEU}${}^\baseline$               \cite{papineni-etal-2002-bleu}                            &       0.941  &       0.891  &       0.864   \\
\nonen \metric{CDER}${}^\baseline$               \cite{leusch-etal-2006-cder}                              &       0.864  &  {\bf 0.949}  &       0.852  \\
\nonen \metric{CharacTER}${}^\submission$          \cite{wang-etal-2016-character}                         &       0.965  &       0.928  &       0.849   \\
\nonen \metric{chrF}${}^\baseline$               \cite{popovic-2015-chrf}                                  &  {\bf 0.974} &       0.931  &       0.864   \\
\nonen \metric{chrF+}${}^\baseline$              \cite{popovic-2017-chrf}                                  &       0.972  &       0.936  &       0.848   \\
\nonen \metric{EED}${}^\submission$                \cite{stanchev-etal-2019-eed}                           &  {\bf 0.982} &  {\bf 0.940}  &       0.851  \\
\nonen \metric{ESIM}${}^\submission$               \cite{chen-etal-2017-enhanced,mathur-etal-2019-putting} &  {\bf 0.980} &  {\bf 0.950}  &  {\bf 0.942} \\
\nonen \metric{hLEPORa\_baseline}${}^\submission$   \cite{han-etal-2012-lepor,han13language}               &       0.941  &       0.814  &   $-$         \\
\nonen \metric{hLEPORb\_baseline}${}^\submission$   \cite{han-etal-2012-lepor,han13language}               &       0.959  &       0.814  &       0.862   \\
\nonen \metric{NIST}${}^\baseline$               \cite{doddington2002automatic}                            &       0.954  &  {\bf 0.916}  & {\bf 0.899}  \\
\nonen \metric{PER}${}^\baseline$                                                                          &      0.875   &      0.857   &       0.869  \\
\nonen \metric{sacreBLEU.BLEU}${}^\baseline$     \cite{post-2018-call}                                     &       0.869  &       0.891  &       0.882   \\
\nonen \metric{sacreBLEU.chrF}${}^\baseline$     \cite{post-2018-call}                                     &  {\bf 0.975} &  {\bf 0.952}  &  {\bf 0.895} \\
\nonen \metric{TER}${}^\baseline$                \cite{snover2006study}                                    &       0.890  &  {\bf 0.956}  & {\bf 0.894}  \\
\nonen \metric{WER}${}^\baseline$                                                                          &      0.872   & {\bf 0.956}  &       0.820  \\
\nonen \metric{YiSi-0}${}^\submission$             \cite{lo-2019-yisi}                                     &  {\bf 0.978} &  {\bf 0.952}  &  {\bf 0.908} \\
\nonen \metric{YiSi-1}${}^\submission$             \cite{lo-2019-yisi}                                     &       0.973  &  {\bf 0.969}  &  {\bf 0.912} \\
\midrule
\nonen \BJTwikiplussysref                                                                                  &      0.976  &      0.936   & {\bf 0.911} \\
\nonen \BJTLASERrefLM                                                                                      & {\bf 0.990} & {\bf 0.935}  & {\bf 0.924} \\
\nonen \BJTmBARTsysref                                                                                    &      0.964    & {\bf 0.944}   &      0.874 \\
\bottomrule
\end{tabular}\caption[WMT19 System-level Results, non-English]{WMT19 System-level results, non-English. 
n denotes number of MT systems. \BoldDenotes{ }
$\baseline$:WMT19 Baseline \cite{ma-etal-2019-results} 
$\submission$:WMT19 Metric Submission \cite{ma-etal-2019-results} 
}\label{fig:mt_metric_wmt19_System_nonEnglish}
\end{table*}

\begin{table*}[h!]
\centering
\footnotesize
    \addtolength{\tabcolsep}{-2pt} 
\begin{tabular}{lrrrrrrrrrrrrrrrrrrrrrrrrrrrrrrrrrrrrrrrrrrrrrrrrr}
\toprule
                                                                                                           & {\bf de--en}     & {\bf fi--en} & {\bf gu--en}    & {\bf kk--en}    & {\bf lt--en} & {\bf ru--en}    & {\bf zh--en} \\
n                                                                                                          &         16      &         12  &         11     &         11     &         11  &         14     &         15  \\
\midrule
\nonen \metric{ibm1-morpheme}${}^\qeasmetric$      \cite{popovic-etal-2011-evaluation}                     &       -0.345  &      0.740  &         $-$    &         $-$    &      0.487  &         $-$    &         $-$ \\
\nonen \metric{ibm1-pos4gram}${}^\qeasmetric$      \cite{popovic-etal-2011-evaluation}                     &       -0.339  &         $-$ &         $-$    &         $-$    &         $-$ &         $-$    &         $-$ \\
\nonen \metric{LASIM}${}^\qeasmetric$                                                                      &       0.247     &         $-$ &         $-$    &         $-$    &         $-$ &    -0.310     &         $-$ \\
\nonen \metric{LP.1}${}^\qeasmetric$                                                                       &       -0.474  &         $-$ &         $-$    &         $-$    &         $-$ &   -0.488     &         $-$ \\
\nonen \metric{UNI}${}^\qeasmetric$                \cite{yankovskaya-etal-2019-quality}                    &  {\bf 0.846}    & {\bf 0.930} &         $-$    &         $-$    &         $-$ & {\bf 0.805}    &         $-$ \\
\nonen \metric{UNI+}${}^\qeasmetric$               \cite{yankovskaya-etal-2019-quality}                    &  {\bf 0.850}    & {\bf 0.924} &         $-$    &         $-$    &         $-$ & {\bf 0.808}    &         $-$ \\
\nonen \metric{YiSi-2}${}^\qeasmetric$             \cite{lo-2019-yisi}                                     &  {\bf 0.796}    &      0.642  & {\bf -0.566} &   -0.324     &      0.442  &   -0.339     & {\bf 0.940} \\
\nonen \metric{YiSi-2\_srl}${}^\qeasmetric$        \cite{lo-2019-yisi}                                     &  {\bf 0.804}    &         $-$ &         $-$    &         $-$    &         $-$ &         $-$    & {\bf 0.947} \\
\midrule
\nonen \BJTwikiplussysgivensrc                                                                             & {\bf 0.890}    & {\bf 0.941} & {\bf 0.171}    & {\bf 0.961} & {\bf 0.989} & {\bf 0.845} & {\bf 0.971} \\
\bottomrule
\end{tabular}\caption[WMT19 System-level Results, to English]{WMT19 System-level results, QE as a metric, to English. n denotes number of MT systems. \BoldDenotes{ }
  $\qeasmetric$:WMT19 QE-as-Metric Submission \cite{fonseca-etal-2019-findings}}\label{fig:mt_metric_wmt19_System_toEnglish-qe}
\end{table*}

\begin{table*}[h!]
\centering
\footnotesize
    \addtolength{\tabcolsep}{-2pt} 
\begin{tabular}{lrrrrrrrrrrrrrrrrrrrrrrrrrrrrrrrrrrrrrrrrrrrrrrrrr}
\toprule
                                                                                                           & {\bf en--cs}     & {\bf en--de}    & {\bf en--fi} & {\bf en--gu} & {\bf en--kk} & {\bf en--lt}   & {\bf en--ru}    & {\bf en--zh} \\
n                                                                                                          &         11      &         22     &         12  &         11  &         11  &         12    &         12     &         12  \\
\midrule
\nonen \metric{ibm1-morpheme}${}^\qeasmetric$      \cite{popovic-etal-2011-evaluation}                     &  {\bf -0.871} &      0.870  &      0.084  &         $-$ &         $-$ & {\bf -0.81} &            $-$ &         $-$  \\
\nonen \metric{ibm1-pos4gram}${}^\qeasmetric$      \cite{popovic-etal-2011-evaluation}                     &          $-$    &      0.393  &         $-$ &         $-$ &         $-$ &           $-$ &            $-$ &         $-$  \\
\nonen \metric{LASIM}${}^\qeasmetric$                                                                      &          $-$    &      0.871  &         $-$ &         $-$ &         $-$ &           $-$ & {\bf -0.823} &         $-$  \\
\nonen \metric{LP.1}${}^\qeasmetric$                                                                       &          $-$    &   -0.569  &         $-$ &         $-$ &         $-$ &           $-$ &      -0.661  &         $-$  \\
\nonen \metric{UNI}${}^\qeasmetric$                \cite{yankovskaya-etal-2019-quality}                    &       0.028     &      0.841  & {\bf 0.907} &         $-$ &         $-$ &           $-$ &    {\bf 0.919} &         $-$  \\
\nonen \metric{UNI+}${}^\qeasmetric$               \cite{yankovskaya-etal-2019-quality}                    &          $-$    &         $-$ &         $-$ &         $-$ &         $-$ &           $-$ &    {\bf 0.918} &         $-$  \\
\nonen \metric{USFD}${}^\qeasmetric$               \cite{ive-etal-2018-deepquest}                          &          $-$    &   -0.224  &         $-$ &         $-$ &         $-$ &           $-$ &    {\bf 0.857} &         $-$  \\
\nonen \metric{USFD-TL}${}^\qeasmetric$            \cite{ive-etal-2018-deepquest}                          &          $-$    &   -0.091  &         $-$ &         $-$ &         $-$ &           $-$ &         0.771  &         $-$  \\
\nonen \metric{YiSi-2}${}^\qeasmetric$             \cite{lo-2019-yisi}                                     &       0.324     &      0.924  &      0.696  & {\bf 0.314} &      0.339  &        0.055  &      -0.766  &   -0.097   \\
\nonen \metric{YiSi-2\_srl}${}^\qeasmetric$        \cite{lo-2019-yisi}                                     &          $-$    &      0.936  &         $-$ &         $-$ &         $-$ &           $-$ &            $-$ &   -0.118   \\
\midrule
\nonen \BJTwikiplussysgivensrc                                                                             & {\bf 0.865}    & {\bf 0.976} & {\bf 0.933} & {\bf 0.444} & {\bf 0.959} &   {\bf 0.908} &    {\bf 0.822} & {\bf 0.793}    \\
\bottomrule
\end{tabular}\caption[WMT19 System-level Results, QE as Metric, from English]{WMT19 System-level results, QE as a metric, from English. n denotes number of MT systems. \BoldDenotes{ } $\qeasmetric$:WMT19 QE-as-Metric Submission \cite{fonseca-etal-2019-findings}}\label{fig:mt_metric_wmt19_System_fromEnglish-qe}
\end{table*}

\begin{table*}[h!]
\centering
\footnotesize
    \addtolength{\tabcolsep}{-2pt} 
\begin{tabular}{lrrrrrrrrrrrrrrrrrrrrrrrrrrrrrrrrrrrrrrrrrrrrrrrrr}
\toprule
                                                                                                           & {\bf de--cs}  & {\bf de--fr}   & {\bf fr--de}  \\
n                                                                                                          &         11   &         11    &         10   \\
\midrule
\nonen \metric{ibm1-morpheme}${}^\qeasmetric$      \cite{popovic-etal-2011-evaluation}                     &      0.355   &    -0.509   & {\bf -0.625} \\
\nonen \metric{ibm1-pos4gram}${}^\qeasmetric$      \cite{popovic-etal-2011-evaluation}                     &         $-$  &       0.085   & {\bf -0.478} \\
\nonen \metric{YiSi-2}${}^\qeasmetric$             \cite{lo-2019-yisi}                                     &      0.606   & {\bf 0.721}   & {\bf -0.53}  \\
\midrule
\nonen \BJTwikiplussysgivensrc                                                                             & {\bf 0.973}  & {\bf 0.889}   & {\bf 0.739}    \\
\bottomrule
\end{tabular}\caption[WMT19 System-level Results, QE as Metric, non-English]{WMT19 System-level results, QE as a metric, non-English. n denotes number of MT systems. \BoldDenotes{ }
  $\qeasmetric$:WMT19 QE-as-Metric Submission \cite{fonseca-etal-2019-findings}}\label{fig:mt_metric_wmt19_System_nonEnglish-qe}
\end{table*}

\end{document}